\pdfoutput=1
\documentclass[11pt]{article}

\usepackage[]{EMNLP2023}

\usepackage{times}
\usepackage{latexsym}

\usepackage{bm}
\usepackage{amsmath}
\usepackage{booktabs}
\usepackage{graphicx}
\usepackage[normalem]{ulem}
\useunder{\uline}{\ul}{}
\usepackage{comment}

\usepackage{amssymb}
\usepackage[T1]{fontenc}
\usepackage[utf8]{inputenc}

\usepackage{microtype}

\usepackage{inconsolata}
\usepackage{enumitem}

\usepackage[nameinlink]{cleveref}
\Crefname{figure}{Figure}{Figures}
\crefname{figure}{Figure}{Figures}
\crefname{example}{Example}{Example}
\crefname{theorem}{Theorem}{Theorem}
\crefname{corollary}{Corollary}{Corollary}
\crefname{lemma}{Lemma}{Lemma}
\crefname{proposition}{Proposition}{Proposition}
\crefname{assumption}{Assumption}{Assumption}
\crefname{section}{Section}{Section}
\crefname{algorithm}{Algorithm}{Algorithm}

\title{FuxiTranyu: A Multilingual Large Language Model \\Trained with Balanced Data}

\author{Haoran Sun, Renren Jin, Shaoyang Xu, Leiyu Pan, Supryadi, \\ {\bf Menglong Cui, } {\bf Jiangcun Du, } {\bf Yikun Lei, } {\bf Lei Yang, } \\ {\bf Ling Shi, } {\bf Juesi Xiao, } {\bf Shaolin Zhu} {\bf and Deyi Xiong${^*}$} \\
  TJUNLP Lab, College of Intelligence and Computing, Tianjin University \\
  \texttt{\{hrsun,rrjin,dyxiong\}@tju.edu.cn}}
\begin{document}
\maketitle

\renewcommand{\thefootnote}{\fnsymbol{footnote}}
\footnotetext[1]{Corresponding author. }
\renewcommand{\thefootnote}{\arabic{footnote}}

\begin{abstract}
    Large language models (LLMs) have demonstrated prowess in a wide range of tasks. However, many LLMs exhibit significant performance discrepancies between high- and low-resource languages.
    To mitigate this challenge, we present \textbf{FuxiTranyu}, an open-source multilingual LLM, which is designed to satisfy the need of the research community for balanced and high-performing multilingual capabilities.
    The base model, FuxiTranyu-8B, features 8 billion parameters and is trained from scratch on meticulously balanced multilingual data that contains 600 billion tokens covering 43 natural languages and 16 programming languages.
    We also develop two instruction-tuned models: FuxiTranyu-8B-SFT which is fine-tuned on a diverse multilingual instruction dataset, and FuxiTranyu-8B-DPO which is further refined with DPO on a preference dataset for enhanced alignment ability.
    Extensive experiments on a wide range of multilingual benchmarks demonstrate the competitive performance of FuxiTranyu against existing multilingual LLMs, e.g., BLOOM-7B, PolyLM-13B, and Mistral-7B-Instruct. 
    Both neuron and representation interpretability analyses reveal that FuxiTranyu achieves consistent multilingual representations across languages. To promote further research into multilingual LLMs, we release both the base and instruction-tuned FuxiTranyu models together with 58 pre-training checkpoints at HuggingFace\footnote{\url{https://huggingface.co/TJUNLP/FuxiTranyu-8B}} and Github.\footnote{\url{https://github.com/tjunlp-lab/FuxiTranyu}}
\end{abstract}

\section{Introduction}
\label{sec:intro}
A well-pretrained base model is crucial for facilitating research and applications of large language models. 
However, training a base LLM from scratch typically demands a substantial amount of data and significant computational resources, posing a barrier to the development of new LLMs. The majority of LLMs are usually tailored to specific languages such as English \cite{llama,llama2} or Chinese \cite{qwen}, neglecting the growing demand for multilingual capabilities, especially from low-resource languages. 
While certain LLMs like Mistral models \cite{mistral} demonstrate multilingual capabilities, their coverage is limited, restricting the exploration in massively multilingual settings.

\begin{table*}[h]
\centering
\small
\scalebox{0.8}{
\begin{tabular}{lllll}
\toprule
LLMs    & Pre-training Tokens & Languages    & \begin{tabular}[c]{@{}l@{}}Base \\ Model Available\end{tabular} & \begin{tabular}[c]{@{}l@{}}Pretraining \\ Checkpoints Available\end{tabular}  \\
\midrule
BLOOM-7B1 \cite{bloom}      & 300B      & 46 NLs + 13 PLs & \checkmark                          & \checkmark              \\
Aya 23-8B \cite{aya23}     & Unknown       & 23 NLs        & $\times$                          & $\times$            \\
PolyLM-13B \cite{polylm}     & 638B      & 18 NLs        & \checkmark                          & $\times$            \\
FuxiTranyu-8B  & 606B      & 43 NLs + 16 PLs & \checkmark                          & \checkmark \\
\bottomrule
\end{tabular}
}
\caption{Comparison between trending multilingual large language models and FuxiTranyu, where NL stands for natural language while PL for programming language.}
\label{model_comparison}
\end{table*}

Recent efforts have been dedicated towards mitigating such language-specific constraints through supervised fine-tuning, as exemplified by Okapi \cite{okapi}. However, as highlighted by the alignment hypothesis in LIMA \cite{lima}, the knowledge of LLMs is predominantly derived from pre-training, while supervised fine-tuning primarily aligns model behavior to instructions, which is a narrow subset of the pre-training data. This makes fine-tuning less effective for boosting multilingual abilities when pre-training is dominated by a few languages.

Other initiatives have focused on pre-training multilingual LLMs, such as BLOOM \cite{bloom} and PolyLM \cite{polylm}. Nevertheless, these efforts are hindered by their performance, which does not measure up to that of current trending LLMs. BLOOM suffers from outdated training data, while PolyLM is undermined by imbalanced language distribution, with English data accounting for approximately 70\% and Chinese for \textasciitilde20\%, potentially leading to insufficient learning of under-represented languages. Previous studies~\cite{xu2024exploring} disclose three traits of multilingual LLMs caused by imbalanced language resources: cross-lingual inconsistency, distorted linguistic relationships, and unidirectional transfer between high- and low-resource languages, emphasizing the need for balanced data distribution.

Recently introduced multilingual LLMs, e.g., Aya 23 models \cite{aya23}, have demonstrated remarkable performance on multiple multilingual benchmarks. They are derived from the CommandR series of models\footnote{\url{https://cohere.com/command}} by performing supervised fine-tuning. However, only the weights of Aya 23 have been released, with its base model remaining undisclosed.

In this work, we present \textbf{FuxiTranyu}, a family of multilingual LLMs supporting 43 natural languages and 16 programming languages. The FuxiTranyu initiative aims to mitigate the aforementioned challenges of multilingual LLMs. The base model comprises 8 billion parameters and has been trained from scratch using approximately 600 billion multilingual tokens. To ensure balanced learning across all supported languages, we have manually controlled the sampling ratio of pre-training data for different languages, striving for as balanced distribution as possible. In line with our commitment to advancing research in multilingual LLMs, we have also released 58 pre-training checkpoints, resonating with the efforts of LLM360 \cite{llm360}. \Cref{model_comparison} compares FuxiTranyu with currently available multilingual LLMs from different perspectives.

In addition to the base model, we develop two instruction-tuned models: FuxiTranyu-8B-SFT, fine-tuned on a collected high-quality multilingual instruction dataset, and FuxiTranyu-8B-DPO, further tuned on preferences with DPO for enhanced alignment ability.

Our evaluations focus on knowledge, capability and alignment dimensions categorized by \citet{guo2023evaluating}. Evaluation results on multilingual discriminative tasks such as multilingual ARC, HellaSwag, and MMLU \cite{okapi}, XWinograd \cite{bloomz, xwinograd2}, XCOPA \cite{xcopa}, XStoryCloze \cite{xstorycloze}, and multilingual generative tasks including WMT and IWSLT translation benchmarks \cite{wmt16,iwslt} and XL-Sum summarization benchmark \cite{xlsum}, demonstrate FuxiTranyu's superior performance compared to BLOOM-7B1 and PolyLM-13B, as detailed in \Cref{sec:evaluation}. The instruction-tuned models, FuxiTranyu-8B-SFT and FuxiTranyu-8B-DPO, also outperform Llama-2-Chat-7B, Mistral-7B-Instruct-v0.1, BLOOMZ-7B1, PolyLM-MultiAlpaca-13B on translation and summarization benchmarks.

To further understand the multilingual capabilities of FuxiTranyu models, we have conducted neuron- and representation-level analysis, revealing that FuxiTranyu-8B learns more language-agnostic representations compared to BLOOM-7B1~\cite{bloom}, which can be attributed to the balanced pre-training data. However, languages with extremely limited resources, such as Bengali and Tamil, are allocated with fewer neurons. Additionally, different layers and components of FuxiTranyu-8B handle multilingual text differently, with deep layers being more language-specific and the importance of attention and MLP components varying across layers.

\section{Related Work}
\label{sec:related-work}
Recent advanced LLMs~\cite{llama,llama2,llama3,qwen,qwen2,01ai-yi,mistral,gemma,gemma-2} have excelled in NLP and cross-modal tasks, sparking increased research on multilingual LLMs~\cite{bloom,palm,polylm}, which aim at broader language support. There are three main approaches to building multilingual LLMs: pre-training from scratch, continual pre-training, and post-training (e.g., supervised fine-tuning and reinforcement learning from human feedback).
% Training multilingual LLMs typically involves a multi-stage process, combining different approaches to enhance the model's capabilities across multiple languages, either training a model from random initialization on massive multilingual data (e.g., BLOOM \cite{bloom}, OPT \cite{opt}, PaLM \cite{palm}, LLaMA \cite{llama}) or building upon existing pretrained LLMs to reduce computational cost (e.g., X-Gen \cite{DBLP:conf/emnlp/VuBLCIC22}, FinGPT \cite{DBLP:conf/emnlp/LuukkonenKLEKKG23}, Cabrita \cite{DBLP:journals/corr/abs-2308-11878}, Sabia \cite{DBLP:journals/corr/abs-2403-09887}). 
% While these methods have made significant strides in bridging the gap between high- and low-resource languages, challenges still remain. 

Pre-training from scratch, like PaLM 1\&2~\cite{palm,palm2}, BLOOM~\cite{bloom}, and PolyLM~\cite{polylm},  leverages extensive training corpora from diverse sources, enabling the incorporation of new knowledge. However, pre-training poses a variety of challenges, such as the need for vast computing resources, which can hinder the development of new multilingual LLMs. Additionally, it also suffers from the curse of multilinguality~\cite{xlmr,ernie-code,llama3,gurgurov2024multilingual}, where the performance of individual languages deteriorates as the number of languages increases. On the other hand, continual pre-training, like Aurora-M~\cite{aurora-m}, LLaMAX~\cite{llamax}, is more efficient but risks catastrophic forgetting of previously learned knowledge.

% From-scratch pre-training often struggles with the curse of multilinguality, where adding more languages can lead to performance degradation for low-resource languages. 
% Continual pre-training, while more efficient, suffers from catastrophic forgetting, where models forget previously learned knowledge. 
Supervised fine-tuning (SFT) often leverages multilingual instruction data or incorporates translation tasks to address data scarcity \cite{DBLP:journals/corr/abs-2305-14705,okapi,DBLP:conf/emnlp/WangMAKMNADASPK22}. 
However, both continual pre-training and SFT rely heavily on high-quality, diverse datasets, which are often limited to many languages. 
Reinforcement Learning from Human Feedback (RLHF) is increasingly used to align models with human preferences \cite{shen2023large}. In multilingual LLMs, multilingual RLHF data are used to train multilingual reward models \cite{DBLP:journals/corr/abs-2401-12246}. 
However, RLHF typically relies on human-annotated data, which can be expensive and time-consuming to collect, especially for under-resourced languages. 
While these methods can achieve impressive performance, they can also be computationally expensive and may not generalize well to unseen languages.

% Recent years have witnessed that prominent MLLMs have been developed, each with specific training methodologies and strengths. 
% These include BLOOM (176B parameters, open-source, over 46 languages), LLaMA (65B parameters, efficient architecture), PaLM (540B parameters, wide benchmark success), OPT (175B parameters, open-source), Qwen (14B parameters, strong benchmark performance ), Mistral (7B parameters, open-source, competitive performance ), and Orion-14B (14B parameters, diverse data of 2.5T tokens, data scheduling strategy). 
% While these models have achieved impressive results, future work should focus on addressing the limitations of existing approaches. 
% We strongly suggest that efforts should be made to develop more robust and efficient training methods and strategies that address the curse of multilinguality, mitigate catastrophic forgetting, alleviate data imbalance, and minimize reliance on expensive annotated data, especially for low-resource languages. 

\section{Pretraining}
\label{sec:pretraining}

We elaborate on the sources and domains of our pre-training data and the efforts we have made in the pre-processing stage in \Cref{data_collection}. Next, we discuss the details of our FuxiTranyu architecture in \Cref{architecture}.
We present the strategy we used to determine which languages should be supported by the FuxiTranyu series of models in \Cref{languages}, the details of our tokenizer training in \Cref{tokenizer}, and the pre-training settings in \Cref{training_details}.

\begin{figure}[t]
\begin{center}
\includegraphics[width=1.0\linewidth]{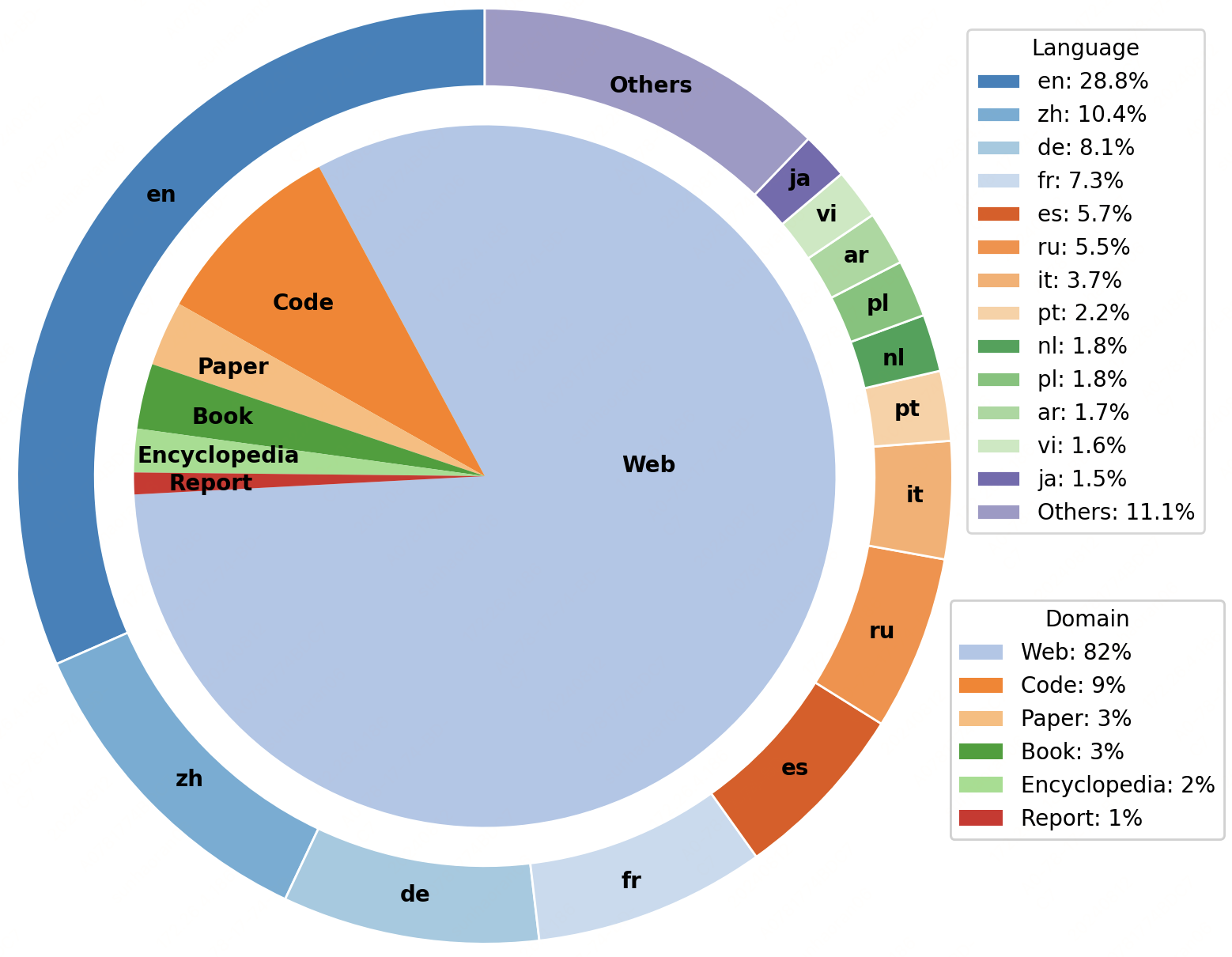} 
\caption{Languages and domains distribution in the pre-training data of FuxiTranyu.}
\label{data-mixture-domain}
\end{center}
\vspace{-5mm}
\end{figure}

\subsection{Data Collection}
\label{data_collection}
The quantity, diversity, and quality of data have proven the most crucial factors determining the performance of a pre-trained base model \cite{chinchilla,llama,llama2}. In pursuit of these objectives, we collect a substantial volume of multilingual data to ensure there are enough tokens for pre-training, in line with scaling laws. Our data collection encompasses a broad spectrum of domains, including public web documents, encyclopedic content, reports, books, scientific articles, and codes. To ensure the quality of the collected corpora, we have employed heuristic quality filters, learned quality filters, and deduplication processes. The composition of the pre-training data mixture is illustrated in \Cref{data-mixture-domain}, and we will delve into the specifics of data collection and pre-processing in the remaining of this section.

A significant portion of our multilingual data comprises web documents, a common approach in open-sourced LLMs \cite{llama,qwen,internlm2,01ai-yi}. We opt to utilize CulturaX \cite{culturax}, a filtered subset of OSCAR \cite{oscar1,oscar2} (itself a subset of Common Crawl) and mC4 \cite{t5} datasets. To improve quality and diversity, we supplement these with data from ROOTS \cite{ROOTS}, MultiUN \cite{multiun1, multiun2}, and OpenSubtitles \cite{opensubtitles2016}, focusing on languages in our language list. Additionally, we incorporate data from encyclopedias, reports, books, and articles, drawing inspiration from Phi series models~\cite{phi1} that achieve strong results using high-quality textbooks. We have collected approximately 500GB of article data from Semantic Scholar (S2ORC) \cite{s2orc}, and around 10GB of Chinese books from the Fudan Cbook dataset.\footnote{\url{https://github.com/FudanNLPLAB/CBook-150K}} We also source multilingual book data from Project Gutenberg, though it forms a small portion of the final corpus. 

Additionally, we collect 535GB of code data from open-source datasets, primarily from Starcoder data,\footnote{\url{https://huggingface.co/datasets/bigcode/starcoderdata}} a subset of the Stack dataset \cite{stack} used to train the StarCoder model \cite{starcoder}. We also include a subset of Github code from the RedPajama dataset.\footnote{\url{https://huggingface.co/datasets/togethercomputer/RedPajama-Data-1T}}

At the filtering stage, we primarily employ three methods similar to prior works \cite{bloom,falcon,qwen,01ai-yi}. The initial filtering phase uses heuristic rules to exclude undesired documents. This involves filtering out documents containing blacklisted URLs or words, such as stop words or flagged words. Subsequently, we filter documents based on statistical information, including the ratio or the number of repeated n-gram characters or words, as well as the document length. Following this, we apply a learned quality filter method based on specific metrics, such as perplexity. In line with the approach taken in BLOOM \cite{bloom}, we utilize KenLM \cite{kenlm} to compute the perplexity of the documents and subsequently filter out those exceeding a predefined threshold.

Upon completion of the quality filter stage, significant efforts are dedicated to data deduplication, as previous studies have emphasized its importance for LLM performance \cite{deduplicating-lee}. We employ MinHash for fuzzy-match deduplication. However, due to the memory-intensive nature of deduplication, processing the entire dataset at once on a server with limited memory is unfeasible. Yet, processing only a portion of the data will not achieve complete deduplication. To address this challenge, we apply a strategy of multi-turn micro-deduplication. We split large documents into chunks and store them in a chunk pool. In each turn, we randomly select chunks from the pool, assemble them back into documents, and perform deduplication on these assembled documents. After processing, the deduplicated documents are again split into chunks and reintegrated into the pool. This process is repeated multiple times until the number of filtered-out documents drops below 1\%. This approach is used for high-resource languages, while low-resource languages are processed in memory due to their smaller dataset size. In the case of code data, we also utilize the MinHash algorithm for data deduplication. Specifically, we leverage the implementation from the bigcode project.\footnote{\url{https://github.com/bigcode-project/bigcode-dataset/blob/main/near_deduplication/minhash_deduplication.py}}

\subsection{Model Architecture}
\label{architecture}
The architecture of FuxiTranyu has been crafted using a modified GPT-2 style framework, drawing inspiration from successful open-source LLMs such as BLOOM, LLaMA, and Qwen. Our modifications are as follows:
\begin{itemize}[leftmargin=*,topsep=0pt,parsep=0pt]
    \item \textbf{Untied Embeddings.} We opt to separate the weights of the input and output embeddings to enhance performance, despite the resulting increase in total model parameters and memory usage.
    \item \textbf{Linear Bias.} In contrast to prior approaches \citep{palm,llama}, we choose not to eliminate the linear bias of the linear projection layers in self-attention and feed-forward layers.
    \item \textbf{Position Encodings.} To extend the model's ability to handle long context, we adopt RoPE \cite{rope}, replacing the original absolute or relative position embedding method utilized in T5 \cite{t5}. RoPE has demonstrated promising results in managing long context situations and has been widely employed in LLMs \cite{llama,baichuan7b,qwen}.
    \item \textbf{Normalization.} Given the significance of pre-training stability in training large LMs with a substantial number of tokens, we implement pre-normalization due to its superior stability compared to post-normalization \cite{pre-norm}. Furthermore, we incorporate the widely used RMSNorm \cite{zhang2019root,rmsnorm} to enhance training efficiency.
    \item \textbf{Activation Function.} While SwiGLU \cite{geglu} has been a popular choice for activation functions due to its performance improvements \cite{scao2022language}, it introduces an additional linear function into the activation process, resulting in a 50\% increase in parameters in the feed-forward layer. Considering this, we decide to use the GeLU \cite{gelu} activation function. GeLU has been shown to achieve similar performance to SwiGLU, as reported in \citep{scao2022language}.
\end{itemize}

\section{Post-training}
\label{sec:alignment}

To develop a model capable of following instructions and engaging in conversational interactions with humans, we have adopted the instruction fine-tuning and reinforcement learning (RL) approach outlined in \citet{instructgpt}.

During the instruction fine-tuning phase, we curate a diverse and high-quality open-source instruction dataset. Given the abundance of instruction-following datasets that have demonstrated exceptional alignment results with various models, manually selecting and fine-tuning the mixture rates for each dataset becomes a challenging task. Consequently, we opt to designate a primary dataset and supplement it with additional datasets. In this context, we select the OpenHermes 2.5 data collection \cite{OpenHermes-2.5} as our base dataset, which is composed of multiple datasets covering a wide range of instructions and yielding excellent results when fine-tuned with Mistral-7B-v0.1. We make modifications to the original OpenHermes 2.5 dataset by replacing Airoboros 2.2 with Airoboros 3.2.\footnote{\url{https://huggingface.co/datasets/jondurbin/airoboros-3.2}} Additionally, we incorporate the Aya dataset \cite{aya_dataset} to enhance the multilingual capabilities of our base model. We filter out the instructions where language is not included in our pre-training language list. To bolster the model's proficiency in Chinese, we include the COIG-CQIA \cite{COIG-CQIA}, ruozhiba-gpt4,\footnote{\url{https://huggingface.co/datasets/hfl/ruozhiba_gpt4}} and in-house Chinese multidisciplinary instruction data as supplementary datasets. To enhance math and coding abilities, we use the dart-math-hard \cite{dart-math} and Magicoder-Evol-Instruct \footnote{\url{https://huggingface.co/datasets/ise-uiuc/Magicoder-Evol-Instruct-110K}}\cite{wizardcoder} datasets. The involved languages in the supervised fine-tuning stage can be found in \Cref{appendix_post_training}.

\begin{table*}[t]
\centering
\footnotesize
\scalebox{0.85}{
\begin{tabular}{ccccccc}
\toprule
Models        &      m-ARC        &        m-Hellaswag  & m-MMLU  & XWinograd & XCOPA & XStoryCloze                  \\ 
& (25-shot) & (10-shot) & (5-xhot) & (5-shot) & (0-shot) & (0-shot)  \\
\midrule
Llama-2-7B    & 35.5 &    48.6  & 35.4     &       78.0  & 58.9 & 55.6 \\
Mistral-7B-v0.1    &  \textbf{40.7}         &        \textbf{54.5} & \textbf{46.7}   &     \textbf{80.5}  & 55.8 & 57.2  \\ 
\midrule
BLOOM-7B1     &     31.8      &   43.4   & 27.1     & 70.0   & 56.9   & 58.2   \\
PolyLM-13B  & 30.6 & 46.0 & 26.4 & 73.4 & 58.9 & 56.4\\
LLaMAX2-7B & 33.1 &  50.3 & 26.7 & 76.9 &  54.5 & 58.8 \\ 
\midrule
FuxiTranyu-8B &     32.7      &  51.8 & 26.6  & 76.1 & \textbf{60.5} & \textbf{58.9}  \\
\bottomrule
\end{tabular}
}
\caption{Average performance of FuxiTranyu-8B base model compared to BLOOM-7B1, PolyLM-13B, Llama-2-7B, Mistral-7B-v0.1, and LLaMAX2-7B on mutlilingual discriminative and generative tasks.}
\label{eval-base}
\end{table*}

In the RL training stage, we opt to use DPO \cite{dpo} as our RL algorithm instead of RLHF \cite{instructgpt,ppo}, as it requires less GPU memory than RLHF, which utilizes PPO as the RL algorithm. We use UltraFeedback \cite{ultrafeedback} for the DPO training, since this dataset focuses on general alignment ability and has been successfully utilized by Zephyr \cite{zephyr} to train the DPO model.

We detail the settings of post-training in \Cref{appendix_post_training}.

\section{Experiments}
\label{sec:evaluation}

\begin{table*}[t]
\centering
\scalebox{0.6}{
\begin{tabular}{cccccccccc}
\toprule
Models        &      m-ARC        &        m-Hellaswag  & m-MMLU  & XWinograd  & XCOPA & XStoryCloze           & Translation & Summarization            \\ 
& (25-shot) & (10-shot) & (5-shot) & (5-shot) & (0-shot) & (0-shot) &(BLEU, 0-shot) & (ROUGE, 0-shot) \\
\midrule
Llama-2-Chat-7B    & 36.4 &       46.3    &  36.0 &      74.8  & 55.9   & 56.5  & 22.1 & 4.6 \\
Mistral-7B-Instruct-v0.1    &  36.3    &    45.5        &   \textbf{39.0}  &   74.0  & 54.5 &   53.4    &    19.1  & 2.2      \\ 
\midrule

BLOOMZ-7B1     &     31.2      &   38.0 & 25.8       & 64.0   & 53.3    &  49.8   &     14.7 & 4.4           \\
PolyLM-MultiAlpaca-13B & 28.6 & 39.1 & 25.9 & 70.9 & 59.9 & 57.0 & - & - \\
LLaMAX2-Alpaca-7B & \textbf{38.7} & \textbf{52.5} & 35.4 & \textbf{77.4} & 56.6  &  \textbf{62.0}  & \textbf{29.1} & 0.3 \\
\midrule
%FuxiTranyu-8B-SFT &     31.8      &   51.5 &  26.8
%& 75.7 & 61.3  &  56.6   & 25.9  & \textbf{8.9}   \\
%FuxiTranyu-8B-DPO &     32.8      &  52.2   &  27.3 & 74.1 & \textbf{62.1}  & 56.9    & 26.4 &  7.3     \\
FuxiTranyu-8B-SFT &     32.8      & 49.2  &  26.9
& 74.7 & 61.2  &  57.4   &  28.3 &   \textbf{9.2} \\
FuxiTranyu-8B-DPO &     34.2      &   47.9  &  27.4 & 69.1 & \textbf{61.8}  & 57.6    & 26.8  &    7.1   \\
\bottomrule
\end{tabular}
}
\caption{Average performance of FuxiTranyu-8B instruct and chat models compared to BLOOMZ-7B1, Llama-2-Chat-7B, and Mistral-7B-Instruct-v0.1 on mutlilingual discriminative and generative tasks.}
\label{eval-instruct}
\vspace{-3mm}
\end{table*}

We conducted extensive experiments to evaluate the capabilities of FuxiTranyu (both the base model and instruction-tuned models) under the multilingual setting. We compared FuxiTranyu against strong baselines, including both English-centric and multilingual models. For English-centric models, we used Llama-2 (Llama-2-7B, Llama-2-chat-7B) \citep{llama2} and Mistral (Mistral-7B-v0.1, Mistral-7B-instruct-v0.1) \citep{mistral} as baseline. For multilingual models, we compared FuxiTranyu with BLOOM (BLOOM-7B1, BLOOMZ-7B1) \citep{bloom,bloomz}, PolyLM (PolyLM-13B, PolyLM-MultiAlpaca-13B) \citep{polylm}, and LLaMAX2 (LLaMAX2-7B, LLaMAX2-7B-Alpaca) \citep{llamax}.\footnote{LLaMAX series models are continual pre-trained on the Llama-2 model to support beyond 100 languages.} We used the LM Evaluation Harness framework \citep{lm-eval-harness} for all evaluation experiments. 

\paragraph{Discriminative Tasks} For evaluating discriminative tasks, we used ARC \citep{arc}, Hellaswag \citep{hellaswag}, MMLU \citep{mmlu}, XWinograd \citep{xwinograd2}, XCOPA \citep{xcopa}, and XStoryCloze \citep{xstorycloze} datasets. Specifically for the multilingual evaluation, we utilized the multilingual version of ARC, HellaSwag and MMLU datasets \citep{okapi} and selected 15 languages for the evaluation (ar, bn, de, en, es, fr, hu, id, it, pt, ru, sk, ta, vi, zh). For XWinograd, XCOPA, and XStoryCloze datasets, we utilized all of the languages provided in the datasets.

\paragraph{Generative Tasks} We evaluated the performance towards generative tasks, especially in translation and summarization tasks. For the translation task, we employed WMT14 in en-fr translation direction \citep{wmt14}, WMT16 in en-de and en-ro translation directions \citep{wmt16} and IWSLT 2017 \citep{iwslt} in en-ar translation direction for measuring the translation performance in our models and benchmark models. For the summarization task, we used XL-Sum \citep{xlsum} dataset. We selected 15 languages for the evaluation (ar, en, es, fr, gu, hi, id, mr, pt, ru, sr, ta, uk, vi, zh).

\subsection{Base Model Evaluation}
First, we report the experiment results of our base models vs. baseline models. We focus on evaluating the capabilities of LLMs towards discriminative tasks. Evaluation results are shown in \Cref{eval-base}. Our model achieves the best performance on the XCOPA and XStoryCloze tasks. For other tasks, our model is significantly better than multilingual models like BLOOM-7B and PolyLM-13B. When compared to LLaMAX2-7B, the evaluation results of our model are almost comparable, with no significant difference from the evaluation results of LLaMAX2-7B. But compared with English-centric models, our model is still worse than Llama-2-7B and Mistral-7B-v0.1 due to the limited training data used for English.

\subsection{Instruction-Tuned Model Evaluation}

%We further compared our instruction-tuned models with other instruction-tuned models. We evaluated these models on both discriminative and generative tasks. Results are shown in \Cref{eval-instruct}. On discriminative tasks, our models achieve the best result on XCOPA. For XWinograd, and XStoryCloze, our models outperform the English-centric models but slightly underperform the multilingual models compared with LLaMAX2-7B. Our models still underperform in m-ARC task due to the limited training data used.
We further compared our instruction-tuned models with other instruction-tuned models. We evaluated these models on both discriminative and generative tasks. Results are shown in \Cref{eval-instruct}. On discriminative tasks, our models achieve the best result on XCOPA. For m-Hellaswag, XWinograd, and XStoryCloze, our models outperform the English-centric models but slightly underperform the multilingual models compared with LLaMAX2-7B. Our models still underperform in m-ARC and m-MMLU tasks due to the limited training data used.

In generative tasks, our models excel on the summarization task, outperforming all baseline models. For the translation task, our models outperform the English-centric models but slightly underperform the multilingual model like LLaMAX2-Alpaca-7B.

More details of our evaluations are discussed in \Cref{appendix-evaluation}, where we report the results for each language tested.
\section{Analysis and Interpretability}
\label{sec:analysis}

We further conducted an interpretability analysis of FuxiTranyu to provide a deep understanding of the underlying mechanisms driving its multilingual capabilities. To ensure a comprehensive analysis and consistency with prior research, we investigated our models from both the neuron~\cite{analysis_xw,analysis_sd,analysis_yq,analysis2,analysis3,analysis4,analysis5} and representation~\cite{analysis10,analysis9,analysis6,analysis7,analysis_sy,analysis_wl,analysis8} perspectives. Specifically, our neuron analysis explores the importance of different neurons to the multilingual abilities of the model, while the representation analysis examines the characteristics of multilingual representations learned by the model. Here, we first introduce the details and results of our neuron analysis, while the representation analysis is discussed in \Cref{sec:representation}.

\subsection{Neuron Analysis}

Neurons in a neural network are the basic computational units of the model. Different inputs may fire neurons in different regions, leading to varied outputs. This computational process can be understood from another perspective: different sets of neurons in the model hold varying degrees of importance for the inputs, thus producing different responses and outputs. To better understand why models generate specific outputs for specific inputs in a multilingual context, we aim to reveal the model's internal mechanisms by evaluating the importance of neurons. Specifically, we assess the importance of different neurons for various linguistic inputs to determine which neurons play a key role in processing particular languages.

We draw on the approach of assessing parameter sensitivity in model pruning, where the basic idea is that a parameter is considered sensitive or important if removing it, by setting the representation produced by that parameter to zero, significantly affects the loss function \cite{analysis2}. Specifically, the model can be represented as a parameter set $\bm{\theta} = [\bm{\theta}_1, \bm{\theta}_2, \dots , \bm{\theta}_n]$, where $\bm{\theta}_i \in \mathbb{R}^d$ is the $i$-th neuron in the model. Let $\bm{h_i}$ denote the representation produced by neuron $\bm{\theta}_i$. The importance of neuron $\bm{\theta}_i$, denoted as $\Phi (i)$, is defined as the change in the loss function $\mathcal{L}$ before and after setting representation $\textbf{h}_{i}$ to zero. Formally, $\Phi (i)$ can be estimated as follows:

\begin{equation}
    \Phi(i) = |\Delta{\mathcal{L}(\textbf{h}_{i})}| = \left|\mathcal{L}\left(\textbf{H}, \textbf{h}_{i}=\textbf{0}\right)-\mathcal{L}\left(\textbf{H}, \textbf{h}_{i}\right)\right|
\label{eq1}
\end{equation}where $\textbf{H}$ is the representation produced by a neuron other than $\bm{\theta}_i$ in the same structure as the $\bm{\theta}_i$.

Calculating the importance of each neuron in the model using the aforementioned method is very time-consuming, as it requires traversing each neuron. However, based on prior studies, we can simplify these calculations using a Taylor expansion, as shown in \Cref{eq2}:

\begin{equation}
\begin{split}
    \Phi(i) & = |\mathcal{L}(\textbf{H}, \textbf{h}_{i}=\textbf{0}) - (\mathcal{L}(\textbf{H}, \textbf{h}_{i}=\textbf{0})\\&+\frac{\partial \mathcal{L}(\textbf{H}, \textbf{h}_{i})}{\partial \textbf{h}_{i}} \textbf{h}_{i}+R_{1}(\textbf{h}_{i}))|
\end{split}
\label{eq2}
\end{equation}

After ignoring the term $R_{1}(\textbf{h}_{i})$, the neuron importance evaluation function is simplified to $\frac{\partial \mathcal{L}(\textbf{H}, \textbf{h}_{i})}{\partial \textbf{h}_{i}} \textbf{h}_{i}$, which is the product of the gradient and the representation. This enables parallel computation of each neuron's importance.

Furthermore, to measure the significance of a specific parameter set $\bm{\alpha} = [\bm{\theta}_l, \bm{\theta}_{l+1}, \dots , \bm{\theta}_k] \subseteq \bm{\theta}$, we compute the importance of each neuron in the set using the following equation:

\begin{equation}
    \Phi (\bm\alpha)=\sum_{i=l}^{k} \Phi (i)
\label{eq3}
\end{equation}

where $\Phi (\bm\alpha)$ denotes the importance of the parameter set $\bm\alpha$. The set $\bm\alpha$ can represent a component or a layer of the model, with the neuron indices in $\bm\alpha$ generally being continuous.

\subsection{Neuron Analysis Setup}

We chose the Flores-200 dataset~\citep{flores} to evaluate the importance of neurons. By selecting the languages ar, bn, es, fr, id, pt, ta, vi, zh, en, de, hu, it, ru, and sk, we analyzed the significance of different model components and layers in response to various linguistic inputs.

\subsection{Neuron Analysis Results}
\label{sec:neuron_5.1.2}
We analyzed the varying importance of different layers across diverse language inputs, as shown in \Cref{layer} (\Cref{sec:neuron_result}). Our findings indicate that universally, shallow layers exhibit low significance while deep layers demonstrate great importance. Notably, languages such as \textit{bn} and \textit{ta} exhibit a notably diminished importance in deep layers compared to others, aligning with our evaluation results where these languages perform poorly. This discrepancy may stem from their relatively limited representation learning in the pre-training data.

We then analyzed the significance of various components across different language inputs, depicted in \Cref{component} (\Cref{sec:neuron_result}), with 8 components per layer. Our findings mirror previous conclusions: components in shallow layers exhibit low importance, whereas those in deep layers show high significance. Moreover, a more detailed observation reveals that MLP components hold greater importance in shallow layers, whereas attention components are more critical in deep layers.

\section{Conclusion}
\label{sec:conclusion}
In this paper, we have presented FuxiTranyu to address the need for open-source multilingual LLMs. Along with the base model, FuxiTranyu-8B, we also present instruction-tuned models fine-tuned on multilingual supervised fine-tuning and preference data, FuxiTranyu-8B-SFT and FuxiTranyu-8B-DPO. Evaluations on multilingual benchmarks show FuxiTranyu outperforms previous multilingual and monolingual LLMs. Furthermore, interpretability analyses underscore the efficacy of the multilingual capabilities embedded in FuxiTranyu.

\section*{Acknowledgements}
The present research was supported by the National Key Research and Development Program of China (Grant No. 2023YFE0116400). The computing resources used in this project were supported by the Scientific Computing Center of the College of Intelligence and Computing, Tianjin University. We would like to thank the anonymous reviewers for their insightful comments.
\bibliography{anthology,custom}
\bibliographystyle{acl_natbib}
\clearpage
\appendix

\begin{table*}[t]
\centering
\small
\begin{tabular}{lll|lll}
\toprule
ISO-931 & Language  & Language Family & ISO-931 & Language   & Language Family \\
\midrule
ar            & Arabic    & Afro-Asiatic    & ky            & Kyrgyz     & Turkic          \\
bg            & Bulgarian & Indo-European   & lo            & Lao        & Kra-Dai         \\
bn            & Bengali   & Indo-European   & ms            & Malay      & Austronesian    \\
ca            & Catalan   & Indo-European   & my            & Burmese    & Sino-Tibetan    \\
cs            & Czech     & Indo-European   & nl            & Dutch      & Indo-European   \\
de            & German    & Indo-European   & pl            & Polish     & Indo-European   \\
el            & Greek     & Indo-European   & pt            & Portuguese & Indo-European   \\
en            & English   & Indo-European   & ro            & Romanian   & Indo-European   \\
es            & Spanish   & Indo-European   & ru            & Russian    & Indo-European   \\
fa            & Persian   & Indo-European   & sv            & Swedish    & Indo-European   \\
fi            & Finnish   & Uralic          & ta            & Tamil      & Dravidian       \\
fr            & French    & Indo-European   & tg            & Tajik      & Indo-European   \\
he            & Hebrew    & Afro-Asiatic    & th            & Thai       & Kra-Dai         \\
hi            & Hindi     & Indo-European   & tk            & Turkmen    & Turkic          \\
hu            & Hungarian & Indo-European   & tl            & Filipino   & Austronesian    \\
id            & Indonesia & Austronesian    & tr            & Turkish    & Turkic          \\
it            & Italian   & Indo-European   & uk            & Ukrainian  & Indo-European   \\
ja            & Japanese  & Japanic         & ur            & Urdu       & Indo-European   \\
kk            & Kazakh    & Turkic          & uz            & Uzbek      & Turkic          \\
km            & Khmer     & Austroasiatic   & vi            & Vietnamese    & Austroasiatic   \\
ko            & Korean    & Koreanic        & zh            & Chinese    & Sino-Tibetan    \\
ku            & Kurdish   & Indo-European   &               &            &      \\
\bottomrule  
\end{tabular}
\caption{The list of 43 natural languages supported by FuxiTranyu.}
\label{natural-language-list}
\end{table*}
\begin{table*}[t]
    \centering
    \small
    \begin{tabular}{lrr|lrr}
    \toprule
        Language & Size (GB) & Ratio (\%) & Language & Size (GB) & Ratio (\%) \\
        \midrule    
        Java & 96 & 17.94 & Go & 26 & 4.86 \\
        JavaScript & 70 & 13.08 & SQL & 11 & 2.06 \\
        Python & 63 & 11.77 & Rust & 9.1 & 1.70 \\
        PHP & 59 & 11.02 & Ruby & 7.9 & 1.48 \\
        C & 53 & 9.90 & Scala & 5.1 & 0.95 \\
        C++ & 52 & 9.72 &  Lua & 3.0 & 0.56 \\
        C\# & 48 & 8,97 & Assembly & 1.6 & 0.30 \\
        TypeScript & 29 & 5.42 & Visual Basic & 1.5 & 0.28 \\
        \bottomrule
    \end{tabular}
    \caption{The list of 16 programming languages covered in FuxiTranyu, including the sizes and ratios of each language.}
    \label{code-language-list}
\end{table*}

\section{Supported Languages in FuxiTranyu}
\label{languages}

Our language selection strategy primarily stems from two distinct perspectives: the availability of pre-training data and geographical considerations. We initially approach language selection from the perspective of available pre-training data. Given that the majority of our pre-training data is sourced from web documents, e.g., CulturaX, we determine the languages for pre-training FuxiTranyu based on the statistical information derived from CulturaX. We select the top 21 languages based on the number of available tokens in descending order. Subsequently, we manually incorporate Asian languages, encompassing those from Southeast Asia, West Asia, and Central Asia, resulting in a total of 43 languages. The complete list can be found in \Cref{natural-language-list}.

In terms of programming languages, we initially consider all 13 languages included in BLOOM \cite{bloom}, such as Java, JavaScript, and Python. Additionally, we include three programming languages (SQL, Assembly, and Visual Basic) due to their high popularity, as indicated by the TIOBE index.\footnote{\url{https://www.tiobe.com/tiobe-index/}} The complete list of programming languages is provided in \Cref{code-language-list}.

\begin{figure*}[ht]
\begin{center}
\includegraphics[width=0.7\linewidth]{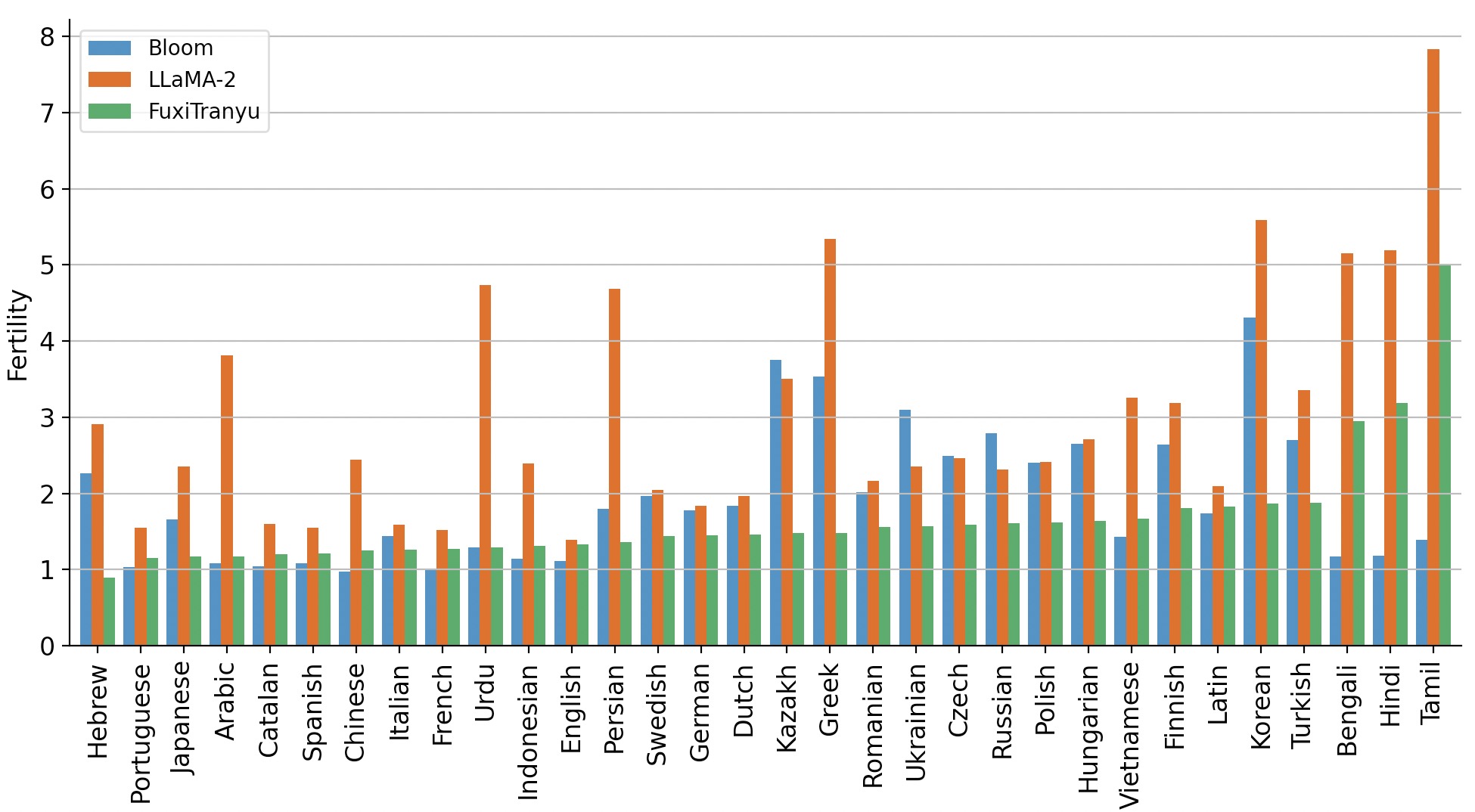} 
\caption{Fertility test results of the tokenizers for FuxiTranyu, Llama-2, and BLOOM.}
\label{fertility}
\end{center}
\end{figure*}

\section{Tokenization}
\label{tokenizer}
We implement the Byte-level Byte-Pair Encoding (BBPE) algorithm using the Hugging Face tokenizer library. Our tokenizer is initiated from GPT-2's tokenizer, incorporating both pre-tokenization and post-tokenization processes. Notably, we opt not to split numbers into digits. In line with the approach outlined in BLOOM \cite{bloom}, we expand the vocabulary size to 250,680 to accommodate multilingual scenarios, thereby mitigating the risk of over-segmentation in low-resource languages.

For training the tokenizer, we randomly sample 1 million documents for each language from our collected data. It's worth noting that for languages with a total document count being less than 1 million, we utilize all available documents in the training data for the tokenizer.

Following the approach used in BLOOM, we also evaluate the performance of our tokenizer using the fertility metric. To assess its efficacy, we conduct a comparative analysis with the Llama-2 and BLOOM tokenizers. This evaluation involves computing fertility on the same set of documents across different languages. Results are presented in \Cref{fertility}, which indicate that the FuxiTranyu tokenizer is more efficient than the others in most languages. Based on our evaluations and interpretability analysis, we believe that the fertility of the tokenizer positively correlates with the model's performance in specific languages. In the fertility test, we observe that Bengali (bn), Hindi (hi), and Tamil (ta) exhibit high fertility, indicating lower tokenization efficiency in these languages compared to others. Consequently, the performance and importance of neurons of these languages in our base model are also suboptimal. Further details are discussed in \Cref{sec:neuron_5.1.2}.

\section{Training Details}
\begin{table*}[]
\centering
\begin{tabular}{ll|ll|ll}
\toprule
\multicolumn{2}{c|}{Pre-Training} & \multicolumn{2}{c|}{SFT} & \multicolumn{2}{c}{DPO} \\
\midrule
\# Params           & 8B    &  Learning Rate & 2e-5 & Learning Rate & 5e-7 \\
Hidden Size         & 4,096  & Warmup Ratio & 10\% & Warmup Ratio & 10\% \\
Intermediate Size   & 16,384 & Batch Size & 512 & Batch Size & 512 \\
Heads               & 32    & Epochs & 2 & Epochs & 1 \\
Layers              & 30    & NEFTune & $\checkmark$ & beta & 0.01 \\
FlashAttn V2    &  $\checkmark$ & FlashAttn V2    &  $\checkmark$ & FlashAttn V2    &  $\checkmark$ \\
Training Tokens & 606B & \# Instances & 1M & \# Instances & 61.1k \\
Position Embed & 4,096   & & & \\
Vocab Size      & 250,752 & & & \\
Learning Rate   & 3e-4 $\rightarrow$ 1e-4 & & & \\
Batch Size      & 2M $\rightarrow$ 4M & & & \\
Context Length  & 4,096  & & & \\
\bottomrule
\end{tabular}
\caption{Model size and hyper-parameters. We append 72 dummy tokens to the vocabulary to make the embedding size be divisible by 128.}
\label{training-details}
\end{table*}

\subsection{Pre-training Details}
\label{training_details}
The training procedure for the FuxiTranyu model adheres to the standard autoregressive language model framework, utilizing the next-token prediction loss as detailed in \citep{gpt3}. To enhance pre-training efficiency, we employ a document packing method similar to that described in \citep{t5}. This involves randomly shuffling documents, merging them, and then truncating them into multilingual chunks that adhere to a maximum context length of 4096 tokens during the pre-training phase.

To mitigate memory consumption and further improve training efficiency, we leverage ZeRO-2 \cite{zero} and Flash-Attention V2 \cite{flashattention-2} technologies. For optimization, the standard AdamW optimizer \cite{adamw} is utilized with hyper-parameters set to $\beta_1=0.9$, $\beta_2=0.95$, and $\epsilon=10^{-8}$. We employ the cosine learning rate scheduler, starting with a maximum learning rate of 3e-4 and decaying to a minimum of 10\% of the maximum rate. Notably, after encountering divergence issues when training approximately 241 billion tokens, we reduced the maximum learning rate to 1e-4 to match with the learning rate used in BLOOM, given the multilingual context of both models. 

Our FuxiTranyu-8B model is trained using the Megatron-LM \cite{megatron} framework on a setup of 32 A800 GPUs, processing a total of 606 billion tokens. The training utilizes FP16 mixed precision to ensure stability. Detailed training parameters and configurations are provided in \Cref{training-details}.

\subsection{Post-Training Details}
\label{appendix_post_training}
The instruction datasets collected do not cover all languages used during pre-training. For the current version of FuxiTranyu, we provide support for the following languages: Arabic, Bengali, Burmese, Chinese, Dutch, English, Filipino, Finnish, French, German, Greek, Hindi, Hungarian, Indonesian, Italian, Japanese, Korean, Kyrgyz, Malay, Persian, Polish, Portuguese, Russian, Spanish, Swedish, Tamil, Thai, Turkish, Ukrainian, Urdu, and Vietnamese.

During the instruction tuning phase, we executed the fine-tuning process on 8 A800 80GB GPUs, leveraging the TRL framework for instruction fine-tuning and DPO training. Throughout both stages, we employed the ChatML format\footnote{\url{https://github.com/openai/openai-python/blob/release-v0.28.0/chatml.md}} for the chat template, and designated \texttt{<PAD>} as the pad token. We used AdamW \cite{adamw} optimizer, complemented by a cosine learning rate scheduler. The maximum sequence length was set to 4096 for both stages.

In the SFT stage, we configured the maximum learning rate to 2e-5, with a warmup phase spanning 10\% of the total steps. The global batch size was set to 512, and the model was trained for 2 epochs. To optimize memory usage, we enabled Flash-Attention V2 \cite{flashattention-2}, ZeRO stage 2 \cite{zero}, and gradient checkpointing. Additionally, we employed NEFTune \cite{neftune}, which introduces noise to embedding weights to enhance the final performance of our instruction-tuned model.

In the subsequent DPO training stage, we adhered to the latest hyper-parameters specified for reproducing the results of Zephyr, as provided by the alignment-handbook.\footnote{\url{alignment_handbook2023}} The beta value for DPO was set to 0.01, and the training took 1 epoch on UltraFeedback. The maximum learning rate was set to 5e-7, with a warmup phase covering 10\% of the total training steps. Similar to the SFT stage, the global batch size was maintained at 512, and we activated Flash-Attention V2 and gradient checkpointing to optimize memory usage. To accommodate the policy and reference model within memory constraints, we utilized ZeRO stage 3 for the policy model and omitted ZeRO for the reference model.

\section{Detailed Evaluation Results}
\label{appendix-evaluation}
We provide detailed evaluation results for each language in this section. First, we present the results for all 15 tested languages on the multilingual ARC in \Cref{arc}, comparing base models and instruction-tuned models. In base models, the results show that our models perform better in 1 of the 15 tested languages for the ARC task. In instruction-tuned models, our models outperforms in ar and vi languages. We speculate that our models still underperformed on this task due to the relatively small amount of training data used.
\begin{table*}[h]
\centering
\begin{tabular}{ccccccccc}
\toprule
Models                   & ar            & bn            & de        &en    & es            & fr            & hu         &id   \\
\midrule
\multicolumn{9}{c}{Base Model}                                                                                           \\
\midrule
Llama-2-7B   & 24.9 & 24.2 & {\ul 37.0} & {\ul 52.5} & {\ul 42.1} & {\ul 43.1} & 31.7 & {\ul 36.1}       \\

Mistral-7B-v0.1   & 30.5 & 23.4 & \textbf{43.1} & \textbf{60.0} & \textbf{52.5} & \textbf{47.7} & \textbf{38.7} & \textbf{39.0}  \\

BLOOM-7B1                & {\ul 31.4}    & 
\textbf{26.2}    & 27.3    & 40.0      & 38.1          & 36.7          & 25.9    &  36.0   \\

PolyLM-13B &  27.3 &	22.4 	&32.8 &	41.8 & 33.2  &	32.7 &	23.6 &	32.8              \\
LLaMAX2-7B &  24.4 &	24.1 &	35.1 & 48.7 &	38.7 & 38.8 &	31.6 &	31.4               \\

FuxiTranyu-8B     &  \textbf{31.5} &  {\ul 25.8} & 36.0 &  38.3 & 35.3 & 35.5 & {\ul 32.0} & 33.3   \\
\midrule
\multicolumn{9}{c}{Instruction-tuned Model}                                                                               \\
\midrule
Llama-2-Chat-7B  & 26.2 & 23.9 &  39.8 & \textbf{53.6} & 43.0 & 42.5 & 32.4  & 35.4   \\

Mistral-7B-Instruct-v0.1  & 23.3  & 24.3 & \textbf{42.5} & 49.7 & {\ul 45.2} & \textbf{46.5} & {\ul 34.1} & 30.0     \\

BLOOMZ-7B1  & 31.2  & 26.2  & 25.4 & 42.7 & 37.2 & 37.6 & 22.8 & 35.9
   \\

PolyLM-MultiAlpaca-13B & 27.4 &	18.4 &	30.5 &		38.2 & 32.9 &	32.8 &	18.6 &	30.2               \\
LLaMAX2-7B-Alpaca &  {\ul 32.4} &	\textbf{27.9} &	{\ul 42.2} &	 {\ul 53.5} & \textbf{45.9}	& {\ul 44.2} &	\textbf{35.6} &	\textbf{38.6}               \\

%FuxiTranyu-8B-SFT  & {\ul 31.7} & {\ul 27.5} & 33.5 & 35.4 & 33.9 & 34.4 & 31.4 & 33.0   \\

%FuxiTranyu-8B-DPO  & \textbf{32.4} & 26.9 & 33.8 & 36.3 & 35.3 & 35.5 & 34.0 & 33.7     \\
FuxiTranyu-8B-SFT  & 31.1 &	26.3 &	33.9 & 38.9	& 35.4 &	 	36.3 &	31.5 &	35.1    \\

FuxiTranyu-8B-DPO  & \textbf{33.3} &	{\ul 27.4} &	35.1 &	39.3 &  38.0 &	 	37.0 &	33.7 &	{\ul 36.9}      \\
\midrule
Models                   & it            & pt            & ru            & sk            & ta            & vi       & zh     \\
\midrule
\multicolumn{8}{c}{Base Model}                                                                                           \\
\midrule

Llama-2-7B   &  {\ul 40.7} & {\ul 41.8} & {\ul 36.9} & 29.5 & 25.0 & 30.7 & 36.2   \\
Mistral-7B-v0.1  & \textbf{49.9}  & \textbf{47.2} & \textbf{42.1} & \textbf{37.1} & \textbf{25.9}  & 31.3 & \textbf{42.8}  \\
BLOOM-7B1   & 29.0 & 38.6 & 27.5 & 24.9 & 24.2 &  \textbf{33.7} & {\ul 37.3}   \\

PolyLM-13B & 32.0 &	34.0 &	32.8 &	23.3 &	{\ul 25.8} &	29.2 &	34.9               \\
LLaMAX2-7B & 36.5 &	37.4 &	33.6 	& {\ul 30.8} &	24.1 &	28.7 &	32.6                \\

FuxiTranyu-8B   &  34.1  & 36.3 & 34.7 & 27.1 & 24.1 & {\ul 32.4} & 34.9     \\
\midrule
\multicolumn{8}{c}{Instruction-tuned Model}                                                                               \\
\midrule

Llama-2-Chat-7B & 41.5 &{\ul 43.3} & 
 \textbf{39.9} & 29.6 & \textbf{26.9} & 31.5 & 37.1   \\

Mistral-7B-Instruct-v0.1 & \textbf{43.3} & \textbf{45.0} & {\ul 39.5} & {\ul 31.1} & {\ul 25.8} & 26.8 & {\ul 37.7}   \\

BLOOMZ-7B1 & 27.5 & 38.7 & 25.5 & 22.5 & 24.2 & {\ul 33.5} & 37.0   \\

PolyLM-MultiAlpaca-13B & 32.6 &	32.7 &	32.5 &	20.3 	& 20.5 &	28.8 &	32.5                \\
LLaMAX2-7B-Alpaca & {\ul 42.8} &	42.7 &	39.4 & \textbf{36.4} &	25.5 &	\textbf{33.7} &	\textbf{39.2}                \\

%FuxiTranyu-8B-SFT & 33.7 & 33.3 & 31.1 & 28.2 & 23.4 & 31.9 & 34.6     \\

%FuxiTranyu-8B-DPO & 34.6 & 34.2 & 32.5 & 29.3 & 
% 24.6 & 32.5  & 36.9  \\

FuxiTranyu-8B-SFT & 33.6 &	35.2 &	32.2 &	29.0 &	23.5 &	32.5 &	36.8      \\

FuxiTranyu-8B-DPO & 36.8 &	37.1 &	33.8 &	29.1 &	25.1 &	\textbf{33.7} &	37.4   \\

\bottomrule
\end{tabular}
\caption{Performance of FuxiTranyu-8B models compared to Llama-2-7B, Mistral-7B-v0.1, BLOOM-7B, PolyLM-13B, and LLaMAX2-7B models on multilingual ARC (25-shot).}
\label{arc}
\end{table*}

Next, we present the results for all 15 tested languages on multilingual HellaSwag in \Cref{hellaswag}, comparing base models and instruction-tuned models. Despite our FuxiTranyu-8B model being trained on only about 600B tokens, it achieves remarkable performance. Comparing base models, our models outperforms other models in ar, bn, hu, and vi languages. Our model still lag behind compared with Mistral-7B-v0.1, but outperform other baseline models, except sk language. The SFT and RL-trained models, FuxiTranyu-8B-SFT and FuxiTranyu-8B-DPO, also deliver promising results across all languages, even competing with powerful monolingual LLMs like Llama-2-7B and Mistral-7B-v0.1, with English and Spanish as an exception.
%But for the SFT and RL-trained models, FuxiTranyu-8B-SFT and FuxiTranyu-8B-DPO, still underperforms other models.
\begin{table*}[h]
\centering
\begin{tabular}{ccccccccc}
\toprule
Models                   & ar            & bn            & de        &en    & es            & fr            & hu         &id   \\
\midrule
\multicolumn{9}{c}{Base Model}                                                                                           \\
\midrule

Llama-2-7B               & 33.7          & 28.7          & 54.0      &  {\ul 78.9}  & 60.4          & 59.1     & 40.7     & 48.5          \\
Mistral-7B-v0.1          & 40.9          & 31.1          & \textbf{61.1} &  \textbf{83.4} &\textbf{67.3} & \textbf{66.5} & {\ul 47.9}       & \textbf{53.2}   \\
BLOOM-7B1                & {\ul 43.3}    & {\ul 32.8}    & 32.4    & 62.1      & 56.7          & 56.6          & 30.1    &   49.5   \\

PolyLM-13B & 39.6 &	28.4 &	49.5 &	71.3 &  55.8 & 	 	54.8 &	29.3 &	50.1  \\
LLaMAX2-7B & {\ul 43.3} &	32.3 &	53.8 &	75.4 & 59.0 &	 	58.1 &	44.1 &	51.0  \\

FuxiTranyu-8B            & \textbf{46.7} & \textbf{33.0} & {\ul 56.2}    &  69.2  &{\ul 60.9}    & {\ul 60.8}    & \textbf{48.2} &  {\ul 52.7} \\
\midrule
\multicolumn{9}{c}{Instruction-tuned Model}                                                                               \\
\midrule
Llama-2-Chat-7B          & 31.4          & 28.3          & 50.7    &  \textbf{78.6}      & {\ul 58.1}          & 57.0          & 39.0    &  44.5      \\
Mistral-7B-Instruct-v0.1 & 31.2          & 28.7          & 52.2      &  70.1   & {\ul 58.1}          &  57.6          & 39.8 & 38.1         \\

BLOOMZ-7B1               &  39.5          & 31.5          & 33.1     &    46.6   & 48.7          & 45.7          & 29.8   & 42.0       \\

PolyLM-MultiAlpaca-13B & 34.0 &	25.7 &	40.7 &	 	66.0 & 43.5 &	43.1 &	26.7 &	40.0 \\
LLaMAX2-7B-Alpaca & 44.7 &	\textbf{33.4} &	\textbf{56.8}	& {\ul 77.3} & \textbf{62.3} &	\textbf{61.4} &	\textbf{45.9} &	\textbf{53.2} \\

%FuxiTranyu-8B-SFT   & {\ul 46.6}    &  32.9    & 56.1  & 69.0   & 60.7    & 61.0    & {\ul 48.2}  & 53.0  \\
%FuxiTranyu-8B-DPO       & \textbf{48.1} & \textbf{33.6} & \textbf{57.7} &  57.8 &\textbf{62.5} & \textbf{62.5} & \textbf{49.3} & \textbf{54.5} \\

FuxiTranyu-8B-SFT   & {\ul 45.1} &	31.9 &	{\ul 53.4} & 64.9 &	57.5 &	 	{\ul 57.9} &	45.1 &	{\ul 49.2}   \\
FuxiTranyu-8B-DPO   &  \textbf{45.2} &	{\ul 33.1} &	51.4 & 57.1 &	55.0 &	 	55.2 &	{\ul 45.5} &	48.7  \\
\midrule
Models                   & it            & pt            & ru            & sk            & ta            & vi       & zh     \\
\midrule
\multicolumn{8}{c}{Base Model}                                                                                           \\
\midrule

Llama-2-7B         & 56.0   &   56.7    & 49.9          & 39.2          & 28.4          & 45.7     & 48.7     \\
Mistral-7B-v0.1  & \textbf{63.0}&  \textbf{65.1}    & \textbf{58.2} & {\ul 46.6} & 29.0          & 47.1   & \textbf{57.2}       \\

BLOOM-7B1            & 40.8     &  56.0   & 32.5          & 29.8          & 29.4    & {\ul 48.3}  & 51.2 \\

PolyLM-13B & 51.4 &	53.7 &	48.7 &	30.1 &	28.0 &	46.8 &	52.0 \\
LLaMAX2-7B & 56.1 &	56.8 &	51.1 &	\textbf{47.8} &	\textbf{30.0} &	47.2 &	49.3 \\

FuxiTranyu-8B   & {\ul 58.4} &  {\ul 59.3}  & {\ul 54.4}    & 43.7    & {\ul 29.9} & \textbf{51.3} & {\ul 52.9} \\
\midrule
\multicolumn{8}{c}{Instruction-tuned Model}                                                                               \\
\midrule
Llama-2-Chat-7B       & 53.7 &     54.0    & 47.6          & 36.4          & 28.8          & 41.2     & 45.1     \\
Mistral-7B-Instruct-v0.1   & 54.6  & 55.8       & 49.6          & 37.4          & 27.7          & 36.1     &  45.9     \\
BLOOMZ-7B1     & 40.3  &    37.3    & 33.1          & 29.6          & 29.5          & 40.6       & 42.6   \\

PolyLM-MultiAlpaca-13B & 40.8 &	42.4 &	40.0 &	27.1 &	25.2 &	38.2 &	\textbf{53.5} \\
LLaMAX2-7B-Alpaca & \textbf{58.7} &	\textbf{59.4} &	\textbf{53.5} &	\textbf{50.3} &	\textbf{30.0} &	\textbf{49.3} &	{\ul 51.9} \\

%FuxiTranyu-8B-SFT  & 57.7 &  59.0  & {\ul 54.0}    & 43.3    & 29.7    & {\ul 50.6} &  51.1   \\
%FuxiTranyu-8B-DPO & \textbf{59.8} & \textbf{60.7}  & \textbf{55.4} & {\ul 44.8} & {\ul 29.9} & \textbf{52.1} & \textbf{54.9} \\
FuxiTranyu-8B-SFT  & {\ul 55.2} &	{\ul 55.9} &	{\ul 51.2} &	{\ul 41.1} &	29.5 &	{\ul 48.7} &	51.3  \\
FuxiTranyu-8B-DPO &  52.9 &	54.7 & 	51.0 &	{\ul 41.1} &	{\ul 29.9} &	{\ul 48.7} &	49.3  \\
\bottomrule
\end{tabular}
\caption{Performance of FuxiTranyu-8B models compared to Llama-2-7B, Mistral-7B-v0.1, BLOOM-7B, PolyLM-13B, and LLaMAX2-7B models on multilingual HellaSwag (10-shot).}
\label{hellaswag}
\end{table*}

We report results on multilingual MMLU in \Cref{mmlu}. Our models still underperform baseline models for all languages. It is in line with the number of training tokens utilized in the pre-training process.
\begin{table*}[h]
\centering
\begin{tabular}{ccccccccc}
\toprule
Models                   & ar            & bn            & de        &en    & es            & fr            & hu         &id   \\
\midrule
\multicolumn{9}{c}{Base Model}                                                                                           \\
\midrule

Llama-2-7B               & {\ul 29.0} &	27.5 &	{\ul 38.8} & 	{\ul 46.0}  & {\ul 39.9}	& {\ul 39.6} &	{\ul 33.3} &	{\ul 37.0}    \\
Mistral-7B-v0.1          &  \textbf{35.8} &	\textbf{32.2} &	\textbf{51.7} &	\textbf{60.7} & \textbf{53.7} &	 	\textbf{53.5} &	\textbf{46.8} &	\textbf{46.9}   \\
BLOOM-7B1                & 27.5 &	{\ul 28.2} & 28.1 & 25.3 & 28.9	&  	27.4 &	26.9 &	26.9  \\
PolyLM-13B & 26.7 &	26.3 &	26.1 & 27.2	& 26.9 &	 	27.2 &	26.4 &	24.9   \\
LLaMAX2-7B & 25.5 &	26.2 &	27.0  & 28.3 &	27.0 	 &	26.7 &	26.9 &	26.8   \\
FuxiTranyu-8B & 26.3 &	25.5 &	27.6 &	27.1 &	27.1 &	27.5 &	26.4 &	26.2          \\
\midrule
\multicolumn{9}{c}{Instruction-tuned Model}                                                                               \\
\midrule
Llama-2-Chat-7B          &  28.5 &	27.0 &	{\ul 39.5} &	{\ul 47.4} &  {\ul 40.8} &	 	{\ul 40.3} &	34.9 &	{\ul 35.8}    \\
Mistral-7B-Instruct-v0.1 & {\ul 29.9} &	{\ul 29.2} &	\textbf{42.2} &	\textbf{51.9} & \textbf{44.3} 	& 	\textbf{44.0} &	{\ul 39.3} &	\textbf{36.5}      \\

BLOOMZ-7B1               &  24.4 &	25.9 &	25.6 &	22.7 & 27.1 &	 	27.7 &	26.1 &	26.3     \\

PolyLM-MultiAlpaca-13B & 25.9 &	26.6 &	26.2 &	 	25.9 & 26.5 & 	26.3 &	25.2 &	25.4    \\
LLaMAX2-7B-Alpaca &  \textbf{30.0} &	\textbf{30.4} &	36.4 &	43.0 &  37.2 &	 	36.9 &	\textbf{47.6} &	35.5   \\

%FuxiTranyu-8B-SFT   & 26.0 &	27.1 &	26.6 &	27.0 &	26.4 &	27.8 &	27.3 &	26.3  \\
%FuxiTranyu-8B-DPO       &  27.0 &	27.3 &	27.2 &	27.0 &	27.4 &	27.8 &	27.6 &	26.4 \\
FuxiTranyu-8B-SFT   & 26.2 &	26.8 &	27.5 &	28.2 &	28.1 &	27.6 &	26.0 &	25.9  \\
FuxiTranyu-8B-DPO       &  27.3 &	27.8 &	27.7 &	28.2 &	27.9 &	27.3 &	26.8 &	26.6 \\
\midrule
Models                   & it            & pt            & ru            & sk            & ta            & vi       & zh     \\
\midrule
\multicolumn{8}{c}{Base Model}                                                                                           \\
\midrule

Llama-2-7B               & {\ul 38.5} &	{\ul 38.7} &	{\ul 35.7} &	{\ul 33.1} &	{\ul 27.2} &	{\ul 32.8} &	{\ul 33.9}   \\
Mistral-7B-v0.1          & \textbf{52.7} &	\textbf{53.4} &	\textbf{49.8} &	\textbf{45.4} &	\textbf{29.7} &	\textbf{41.5} &	\textbf{46.0}    \\
BLOOM-7B1                & 25.7 &	25.3 &	26.2 &	26.1 &	26.6 &	28.1 &	29.1   \\
PolyLM-13B & 27.5 &	24.5 &	26.3 &	27.4 &	26.4 &	25.3 &	26.8   \\
LLaMAX2-7B & 27.0 &	26.9 &	27.0 &	26.6 &	26.2 &	26.8 &	26.1   \\
FuxiTranyu-8B            &  27.1 &	26.8 &	27.7 &	26.0 &	26.3 & 	26.3 &	26.0 \\
\midrule
\multicolumn{8}{c}{Instruction-tuned Model}                                                                               \\
\midrule
Llama-2-Chat-7B          & {\ul 39.7} &	{\ul 40.2} &	{\ul 36.8} &	{\ul 33.7} &	27.0 &	32.7 &	{\ul 35.2}     \\
Mistral-7B-Instruct-v0.1 &  \textbf{42.5} &	\textbf{43.4} &	\textbf{41.6} &	\textbf{37.8} &	{\ul 27.7} &	\textbf{34.0} &	\textbf{40.1}     \\

BLOOMZ-7B1               &  25.8 &	22.8 &	25.4 &	26.3 &	26.7 &	26.3 &	27.2     \\

PolyLM-MultiAlpaca-13B & 25.9 &	26.2 &	26.2 &	25.5 &	25.5 &	25.7 &	26.1    \\
LLaMAX2-7B-Alpaca & 37.5 &	35.7 &	32.6 &	33.0 &	\textbf{28.4} &	{\ul 33.6} &	33.4    \\

%FuxiTranyu-8B-SFT   & 27.1 &	27.0 &	26.8 &	27.2 &	26.4 &	25.9 &	27.0  \\
%FuxiTranyu-8B-DPO       & 27.5 &	27.7 &	28.0 &	27.6 &	26.9 &	26.2 &	27.7  \\
FuxiTranyu-8B-SFT   & 26.2 &	25.9 &	27.9 &	26.6 &	27.0 &	26.4 &	26.8  \\
FuxiTranyu-8B-DPO       & 28.6 &	27.1 &	28.2 &	26.7 &	26.8 &	26.7 &	28.0  \\
\bottomrule
\end{tabular}
\caption{Performance of FuxiTranyu-8B models compared to Llama-2-7B, Mistral-7B-v0.1, BLOOM-7B, PolyLM-13B, and LLaMAX2-7B models on multilingual MMLU (5-shot).}
\label{mmlu}
\end{table*}

Results on XWinograd are depicted in \Cref{xwinograd}. In base models, although our models still underperformed compared to Mistral-7B-v0.1, they outperforms previous multilingual LLMs like BLOOM-7B1 and PolyLM-13B across all languages. Notably, our FuxiTranyu SFT and DPO models achieve better results in Chinese. 
\begin{table*}[t]
\centering
\begin{tabular}{ccccccc}
\toprule
Models                   & fr            & pt            & zh            & en            & ru            & jp            \\
\midrule
\multicolumn{7}{c}{Base}                                                                                                 \\
\midrule
Llama-2-7B               & \textbf{81.9} & 74.9          & 74.4          & {\ul 90.4}    & {\ul 72.1}    & 74.0    \\
Mistral-7B-v0.1          & \textbf{81.9} & \textbf{80.6} & \textbf{80.0} & \textbf{90.6} & \textbf{72.4} & \textbf{77.5} \\
BLOOM-7B1                & 71.1          & 76.8          & 74.4          & 82.2          & 56.8          & 58.5          \\
PolyLM-13B               & 73.5          & 74.9          & 76.6          & 84.6          & 65.1          & 65.7          \\
LLaMAX-7B & 77.1 & 76.8 & 75.4 & 87.8 & 69.8 & {\ul 74.4}    \\
FuxiTranyu-8B            & {\ul 78.3}    & {\ul 77.2}    & {\ul 76.8}    & 85.4          & 66.4          & 72.4          \\
\midrule
\multicolumn{7}{c}{Instruction-tuned Model}                                                                               \\
\midrule
Llama-2-Chat-7B          & {\ul 79.5} & 71.9          & 62.9          & {\ul 88.3}    & 67.6          & {\ul 70.7}          \\
Mistral-7B-Instruct-v0.1 & 77.1    & 71.5          & {\ul 74.0}          & \textbf{89.8} & {\ul 70.5} & 67.5          \\
BLOOMZ-7B1               & 68.7          & 65.4          & 71.0          & 83.5          & 53.7          & 56.4          \\
PolyLM-MultiAlpaca-13B   & 71.1          & 72.2          & 73.6          & 83.9          & 67.9          & 65.2          \\
LLaMAX-7B-Alpaca & \textbf{81.9} & \textbf{76.8} & 72.2 & {\ul 88.3} & \textbf{71.8} & \textbf{73.7}    \\
%FuxiTranyu-8B-SFT   &  77.1    & \textbf{76.8} & {\ul 76.8}    & 85.6          & 68.3    & 73.1    \\
%FuxiTranyu-8B-DPO       & 72.3          & {\ul 74.5}    & \textbf{78.2} & 84.2          & 67.0          & {\ul 73.2} \\
FuxiTranyu-8B-SFT   &   75.9 & {\ul 76.4} & \textbf{75.2} & 83.7 & 68.3 & 68.7   \\
FuxiTranyu-8B-DPO   &   77.1 & 67.3 & 66.7 & 73.9 & 62.9 & 66.5   \\
\bottomrule
\end{tabular}
\caption{Performance of FuxiTranyu-8B models compared to Llama-2-7B, Mistral-7B-v0.1, BLOOM-7B1, PolyLM-13B, and LLaMAX2-7B models on XWinograd (5-shot).}
\label{xwinograd}
\end{table*}

Results on XCOPA and XStoryCloze are shown in \Cref{xcopa} and \Cref{xstory}. For XCOPA, our base models achieve better results in sw, ta, tr, and vi. When compared to instruction-tuned models, our models achieve better results in more languages, specifically in it, ta, th, tr, and vi. On the XStoryCloze task, our base models achieve better results in three languages: ar, my, and ru. However, for instruction-tuned models, our models outperform other baseline models only in my.
\begin{table*}[t]
\centering
\resizebox{\textwidth}{!}{
\begin{tabular}{cccccccccccc}
\toprule
Models                   & et	& ht &	it &	id &	qu &	sw &	ta &	th &	tr &	vi &	zh            \\
\midrule
\multicolumn{12}{c}{Base}                                                                                                 \\
\midrule
Llama-2-7B               &  48.6	& 50.6	& {\ul 65.8} &	62.4 &	\textbf{51.4} &	52.2 &	53.4 &	56.4 &	54.8 &	63.0 &	65.0 \\
Mistral-7B-v0.1          & 47.0 &	{\ul 51.4} &	{\ul 65.8} &	58.2 &	48.6 &	51.2 &	53.8 &	57.0 &	56.8 &	58.8 &	65.2 \\
BLOOM-7B1                &  48.2 &	50.8 &	52.8 &	{\ul 69.8} &	{\ul 50.8} &	51.6 &	{\ul 59.2} &	55.4 &	51.2 &	70.8 &	65.2\\
PolyLM-13B               &  \textbf{49.8}	& 50.4	& \textbf{66.0} &	\textbf{70.2} &	50.4 &	51.8 &	55.0 &	\textbf{58.6} &	{\ul 57.8}	 & {\ul 70.8} &	\textbf{67.0}    \\
LLaMAX-7B &  {\ul 49.2} &	\textbf{52.6} &	52.6 &	53.8 &	\textbf{51.4} &	{\ul 54.0} &	58.0 &	57.2 &	53.0 &	53.0 &	63.4 \\
FuxiTranyu-8B            &  {\ul 49.2} &	51.2 &	71.4	& 69.6 &	49.6 &	\textbf{55.4} &	\textbf{60.0} &	{\ul 58.0} &	\textbf{62.4} &	\textbf{72.8} &	{\ul 65.8} \\
\midrule
\multicolumn{12}{c}{Instruction-tuned Model}                                                                               \\
\midrule
Llama-2-Chat-7B          &  47.8 &	51.4 &	67.0 &	62.4 &	50.8 &	52.2 &	50.6 &	54.8 &	55.6 &	61.6 &	61.2      \\
Mistral-7B-Instruct-v0.1 &  48.2 &	51.2 &	65.4 &	54.0 &	49.2 &	{\ul 54.6} &	55.2 &	53.2 &	52.2 &	53.2 &	63.4   \\
BLOOMZ-7B1               &  49.2 &	51.4 &	51.8 &	58.2 &	{\ul 52.2} &	53.2 &	54.6 &	54.4 &	53.0 &	55.8 &	52.8      \\
PolyLM-MultiAlpaca-13B   & 47.8 &	50.4 &	65.0 &	\textbf{70.0} &	51.0 &	52.4 &	55.6 &	{\ul 59.0} &	59.8 &	73.4 &	\textbf{74.8}      \\
LLaMAX-7B-Alpaca & \textbf{51.2} &	\textbf{54.2} &	61.0 &	57.2 &	\textbf{52.4} &	\textbf{55.0} &	57.0 &	56.4 &	55.4 &	55.4 &	67.6   \\
%FuxiTranyu-8B-SFT   & {\ul 49.6} &	{\ul 53.2} &	{\ul 71.8} &	69.8 &	51.8 &	53.2 &	{\ul 61.0} &	\textbf{61.2} &	{\ul 62.8} &	71.8 &	{\ul 67.8}    \\
%FuxiTranyu-8B-DPO       & 47.4 &	52.6 &	\textbf{73.4} &	\textbf{73.0} &	51.0 &	53.0 &	\textbf{61.8} &	{\ul 59.8} & 	\textbf{63.6} 	& \textbf{76.6} &	\textbf{70.8}  \\
FuxiTranyu-8B-SFT   & 49.4& 	{\ul 51.8} & 	{\ul 71.6}	&66.8 	&50.6 	&53.0 &	{\ul 62.0} &	\textbf{60.8} &	{\ul 63.6} &	{\ul 73.6} &	69.6  \\
FuxiTranyu-8B-DPO   &  {\ul 49.6} &	51.2 &	\textbf{75.6} &	{\ul 69.2} &	48.6 &	52.6 &	\textbf{63.0} &	{\ul 59.0} &	\textbf{65.6} &	\textbf{74.6} &	{\ul 70.6}  \\
\bottomrule
\end{tabular}
}
\caption{Performance of FuxiTranyu-8B models compared to Llama-2-7B, Mistral-7B-v0.1, BLOOM-7B1, PolyLM-13B, and LLaMAX2-7B models on XCOPA (0-shot).}
\label{xcopa}
\end{table*}
\begin{table*}[t]
\centering
\begin{tabular}{ccccccccccc}
\toprule
Models                   & ar  &	es &	eu	& hi &	id &	my &	ru &	sw	 & te &	zh    \\
\midrule
\multicolumn{11}{c}{Base}                                                                                                 \\
\midrule
Llama-2-7B               & 49.6 & {\ul 67.4}	& 50.4 &	53.7 &	59.3 &	48.1 &	62.9 &	50.5 &	54.3 &	59.5  \\
Mistral-7B-v0.1          & 53.1 & \textbf{69.0} &	51.2 &	55.4 &	59.2 &	48.7 &	{\ul 66.7} &	51.6 &	54.1 &	\textbf{63.3}  \\
BLOOM-7B1                & 58.6 & 66.1 &	\textbf{57.2} &	\textbf{60.6} &	\textbf{64.5} &	49.0 &	52.7 &	{\ul 53.9} &	{\ul 57.4} &	61.9 \\
PolyLM-13B               &  56.5 & 65.6 &	51.6 &	48.8 &	{\ul 63.9} &	47.3 &	64.1 &	49.3 &	53.7 &	\textbf{63.3}  \\
LLaMAX2-7B & {\ul 58.8} & 65.3 &	{\ul 54.5} &	58.2 &	60.6 &	{\ul 52.2} &	61.2 &	\textbf{57.2} &	\textbf{59.3} &	60.8\\
FuxiTranyu-8B            & \textbf{59.2} & 66.1 &	52.1 &	{\ul 59.4} &	63.8  &	\textbf{56.9} &	\textbf{67.6} &	49.0 &	52.5 &	{\ul 62.1}\\
\midrule
\multicolumn{11}{c}{Instruction-tuned Model}                                                                               \\
\midrule
Llama-2-Chat-7B          & 50.1  &	{\ul 67.1} &	51.0 &	54.4 &	60.2 &	48.8 &	65.3 &	{\ul 52.1} &	53.7 &	62.4  \\
Mistral-7B-Instruct-v0.1 & 47.1 & 	63.3 &	50.0 &	49.8 &	52.3 &	47.6 &	62.3 &	49.6 &	51.8 &	59.7  \\
BLOOMZ-7B1               & 47.9 & 	51.0 &	48.6 &	50.8 &	51.0 &	47.4 &	46.9 &	 50.4 &	{\ul 54.0} &	50.0    \\
PolyLM-MultiAlpaca-13B  &   57.2 & 	66.0 &	51.2 &	49.0 &	{\ul 65.3} &	47.2 &	{\ul 65.5} &	48.4 &	53.1 &	\textbf{66.8}  \\
LLaMAX2-7B-Alpaca & \textbf{60.4}  &	\textbf{70.6} &	\textbf{54.8} &	\textbf{62.1} &	\textbf{66.5} &	53.8 &	\textbf{67.4} &	\textbf{60.1} &	\textbf{59.3} &	{\ul 65.3}  \\
%FuxiTranyu-8B-SFT   & 57.1 & 	63.5 &	{\ul 51.5} &	56.2 &	59.9 &	53.5 &	62.7 &	49.0 &	53.2 &	59.6 \\
%FuxiTranyu-8B-DPO       & 55.9 & 	63.1 &	51.4 &	{\ul 58.4} &	59.8 &	\textbf{54.9} &	62.2 &	48.1 &	53.1 &	61.8   \\
FuxiTranyu-8B-SFT   & 57.6  & 65.4 &	{\ul 51.4} &	56.8 &	61.5 &	\textbf{54.7} &	63.7 &	49.8 &	53.3 &	59.4  \\
FuxiTranyu-8B-DPO   &  {\ul 60.2}  & 64.9 &	49.8 &	{\ul 58.1} &	62.3 &	{\ul 54.6} &	63.7 &	49.0 &	52.2 &	61.0 \\
\bottomrule
\end{tabular}
\caption{Performance of FuxiTranyu-8B models compared to Llama-2-7B, Mistral-7B-v0.1, BLOOM-7B1, PolyLM-13B, and LLaMAX2-7B models on XStoryCloze (0-shot).}
\label{xstory}
\end{table*}

%We present our evaluation results for generative tasks in Table \ref{xlsum} and Table \ref{translation}. On the XL-Sum task, our models significantly outperform all baseline models across all evaluated languages, demonstrating the potential of our models on summarization task, particularly in a multilingual context. For the translation tasks in WMT14, WMT16, and IWSLT2017, our models excel in the en-ro, en-de, and en-fr translation directions. However, they still lag behind other baseline models in the ro-en, de-en, fr-en, ar-en, and en-ar translation directions. This indicates that our models perform significantly better for out-of-English translation directions. Although our models underperform in the en-ar direction compared to LLaMAX-2-Alpaca, they still achieve notably better results than other models.
We present our evaluation results for generative tasks in \Cref{xlsum} and \Cref{translation}. On the XL-Sum task, our models significantly outperform all baseline models across all evaluated languages, demonstrating the potential of our models on summarization task, particularly in a multilingual context. For the translation tasks in WMT14, WMT16, and IWSLT2017, our models excel in the en-ro and en-de translation directions. However, they still lag behind other baseline models in the ro-en, de-en, fr-en, ar-en, and en-ar translation directions. This indicates that our models perform significantly better for out-of-English translation directions. Although our models underperformed in the en-fr and en-ar directions compared to LLaMAX2-Alpaca, they still achieve notably better results than other models.

\begin{table*}[t]
\centering
\resizebox{\textwidth}{!}{
\begin{tabular}{cccccccccccccccc}
\toprule
Models                   & ar  &	en &	es	 & fr & gu &   hi &	   id &  mr & 	pt &	ru &  sr &	ta &  uk & vi &	zh	 \\
\midrule

%\multicolumn{19}{c}                                                
Llama-2-Chat-7B             & 0.5  & {\ul 11.0}   & 11.0   & 9.8  & 0.5 & 0.2 & 6.1  & 0.2 & 8.9  & 2.8 &  3.2 & 0.8  & {\ul 2.3} & 10.1 & {\ul 1.0}    \\
Mistral-7B-Instruct-v0.1    & 0.1  & {\ul 11.0}   & 3.0    & 3.4  & 0.3 & 0.2 & 3.1  & 0.6 & 3.2  & 0.4 & 2.1 & 0.2  & 0.3 & 4.6  & 0.6  \\
BLOOMZ-7B1                  & 0.3 & 7.6  & {\ul 13.7} & {\ul 13.1} & 0.4 & 0.0   & 1.2  & 0.0   &  {\ul 13.1} & 0   & 1.7 & 0.0     & 0.0   & 15.4 & 0.0    \\
LLaMAX2-7B-Alpaca           & 0.0   & 1.7  & 0.5  & 0.7  & 0.0   & 0.0   & 0.3  & 0.0   &  0.2  & 0.0   & 0.5 & 0.1   & 0.1 & 0.2  & 0.0    \\
%FuxiTranyu-8B-SFT & {\ul 2.0}    & \textbf{13.3} & \textbf{16.3} & \textbf{16.7} & \textbf{0.8} & {\ul 1.5} & \textbf{13.9} & {\ul 1.8} &  \textbf{17.5} & {\ul 6.0}   & \textbf{3.3} & {\ul 1.4}  & {\ul 5.2} & \textbf{28.4} & \textbf{6.1}  \\
%FuxiTranyu-8B-DPO     & \textbf{2.9}  & {\ul 10.3} & 12.5 & 11.4 & {\ul 0.7} & \textbf{2.3} & {\ul 10.4} & \textbf{3.1} & {\ul 13.7} & \textbf{6.5} & 2.0   & \textbf{3.1} & \textbf{5.5} & {\ul 20.1} & {\ul 5.4} \\
FuxiTranyu-8B-SFT & {\ul 1.9} &	\textbf{11.8} &	\textbf{16.3} &	\textbf{16.6} &  {\ul 0.7} &	{\ul 1.6}  & \textbf{17.8}  & {\ul 2.1} &  \textbf{17.5} &	{\ul 6.4} & \textbf{6.1} &  {\ul 1.3} &  \textbf{5.3} & \textbf{27.7} &	\textbf{5.6}  \\
FuxiTranyu-8B-DPO &  \textbf{2.8} &	9.5 &	11.1 &	11.0 & \textbf{0.9} &	\textbf{2.4} & {\ul 10.7} & \textbf{3.2} & 12.3 &	\textbf{6.5} & {\ul 4.0} &   \textbf{2.8} &  \textbf{5.3} & {\ul 18.3} &	\textbf{5.6}    \\

\bottomrule
\end{tabular}
}
\caption{Performance of FuxiTranyu-8B models compared to Llama-2-7B, Mistral-7B-v0.1, BLOOM-7B1, and LLaMAX2-7B models on XL-Sum (0-shot).}
\label{xlsum}
\end{table*}
% \usepackage{multirow}
% \usepackage{graphicx}

\begin{table*}[]
\footnotesize
\begin{tabular}{ccccccccc}
\toprule
\multicolumn{1}{c}{Models} &
  \multicolumn{2}{c}{WMT16 (EN-RO)} &
  \multicolumn{2}{c}{WMT16 (RO-EN)} &
  \multicolumn{2}{c}{WMT16 (EN-DE)} &
  \multicolumn{2}{c}{WMT16 (DE-EN)} \\
\multicolumn{1}{c}{} &
  BLEU &
  \multicolumn{1}{c}{CHRF} &
  BLEU &
  \multicolumn{1}{c}{CHRF} &
  BLEU &
  \multicolumn{1}{c}{CHRF} &
  BLEU &
  \multicolumn{1}{c}{CHRF}  \\ \midrule
Llama-2-Chat-7B             & 17.18 & 44.20 & 31.43 & 58.00 & 20.01 & 48.31 & 35.41 & 60.78 \\
Mistral-7B-Instruct-v0.1     & 13.66 & 41.47 & 24.58 & 53.04 & 19.41 & 49.25 & 30.19 & 58.27  \\

BLOOMZ-7B1                  & 1.88  & 20.09 & 11.35 & 36.22 & 3.76  & 23.27 & 22.30 & 46.69  \\
LLaMAX2-7B-Alpaca &
  24.52 &
  51.94 &
  \textbf{36.02} &
  {\ul 60.85} &
  {\ul 26.31} &
  53.95 &
  \textbf{37.05} &
  \textbf{61.90} 
  \\
%FuxiTranyu-8B-SFT & {\ul 26.29} & {\ul 54.18} & 27.18 & 55.12 & \textbf{27.94} & \textbf{57.75} & 32.99 & 60.00     \\
%FuxiTranyu-8B-DPO & \textbf{26.48} & \textbf{54.94} & 30.69 & {\ul 59.12} & {\ul 26.65} & {\ul 57.43} & 32.15 & 60.26 \\

FuxiTranyu-8B-SFT & \textbf{25.64} &	\textbf{53.07} &	{\ul 34.96} &	\textbf{61.33} &	\textbf{27.03} &	{\ul 56.4} &	{\ul 35.91} &	{\ul 61.55} \\
FuxiTranyu-8B-DPO & {\ul 24.8} &	{\ul 53.06} &	32.9 &	59.97 &	25.57 &	\textbf{56.42} &	33.52 &	60.43 \\
  
  \midrule
  \multicolumn{1}{c}{Models} &
  \multicolumn{2}{c}{WMT14 (EN-FR)} &
  \multicolumn{2}{c}{WMT14 (FR-EN)} &
  \multicolumn{2}{c}{IWSLT2017-AR-EN} &
  \multicolumn{2}{c}{IWSLT2017-EN-AR} \\
\multicolumn{1}{c}{} &
  BLEU &
  \multicolumn{1}{c}{CHRF} &
  BLEU &
  \multicolumn{1}{c}{CHRF} &
  BLEU &
  \multicolumn{1}{c}{CHRF} &
  BLEU &
  \multicolumn{1}{c}{CHRF} \\ \midrule
Llama-2-Chat-7B              & 24.97 & 52.34 & {\ul 34.49} & 60.89 & 12.51 & 36.18 & 1.15 & 17.73 \\
Mistral-7B-Instruct-v0.1     & 24.24 & 52.08 & 31.40 & 59.50 & 9.13  & 32.64 & 0.31 & 13.31 \\

BLOOMZ-7B1                   & 17.73 & 41.02 & 31.07 & 56.03 & 25.25 & 47.64 & 4.58 &  25.05 \\
LLaMAX2-7B-Alpaca &
  \textbf{32.86} &
  59.53 &
  \textbf{36.00} &
  \textbf{61.64} &
  \textbf{29.76} &
  {\ul 52.68} &
  \textbf{10.47} &
  \textbf{40.27} 
  \\
%FuxiTranyu-8B-SFT   & \textbf{34.06} & \textbf{60.74} & 28.83 & 57.86 & 21.42 & 42.91 & 8.19 & 35.67 \\
%FuxiTranyu-8B-DPO & {\ul 33.15} & {\ul 60.66} & 31.02 & 59.82 & 22.83 & {\ul 49.30} & {\ul 8.47} & {\ul 36.82} \\
FuxiTranyu-8B-SFT   & {\ul 32.82} &	{\ul 59.57} &	34.07 &	{\ul 61.1} &	{\ul 28.83} &	\textbf{52.79} &	{\ul 7.15} &	{\ul 31.14} \\
FuxiTranyu-8B-DPO & 31.98 &	\textbf{59.64} &	32.27 &	60.19 &	27.05 &	51.5 &	6.5 &	29.41 \\
  \bottomrule
\end{tabular}
\caption{Performance of FuxiTranyu-8B models compared to Llama-2-7B, Mistral-7B-v0.1, BLOOM-7B1, and LLaMAX2-7B models on WMT14, WMT16, and IWSLT2017 (0-shot).}
\label{translation}

\end{table*}

\section{Additional Analysis and Interpretability}
\begin{figure*}[t]
\begin{center}
\includegraphics[width=0.8\linewidth]{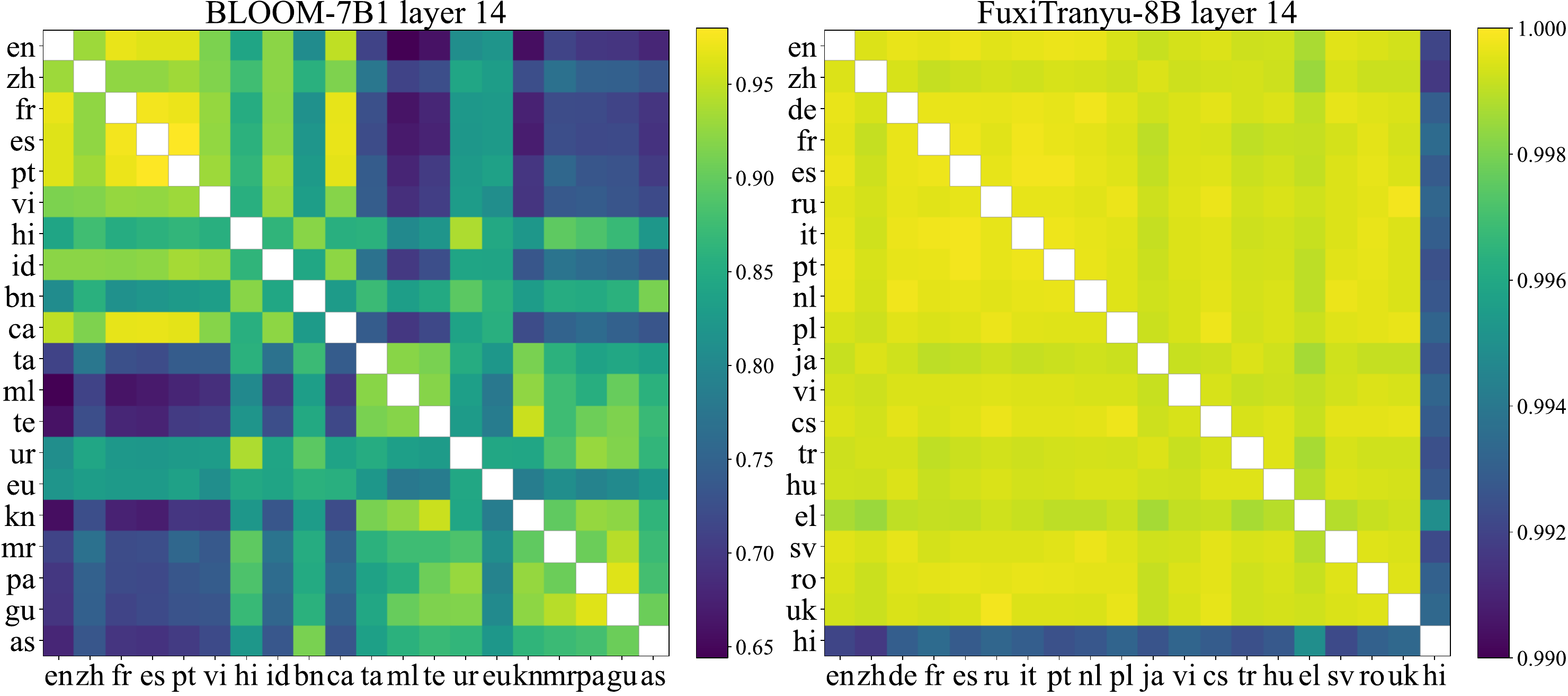} 
\caption{Similarity distribution of multilingual representations in the intermediate layers of BLOOM-7B1 and FuxiTranyu-8B, with languages sorted based on their percentages in the pre-training data.}
\label{similarity}
\end{center}
\end{figure*}

\subsection{Representation Analysis}
\label{sec:representation}

Language models encode textual symbols into high-dimensional representations with rich semantic information. For a multilingual language model, due to parameter sharing mechanisms, it encodes textual symbols from different languages into a unified representation space. Furthermore, through multilingual joint training, the model learns multilingual representations, which encode the intrinsic characteristics of languages and the relationships between different languages. Here, we explore the multilingual characteristics of the model from the perspective of the multilingual representations it learns. Specifically, we calculate the similarity of representations across different languages.

To quantitatively evaluate the similarity between different language representations, we choose cosine similarity for its simplicity and effectiveness. To mitigate the impact of semantic differences on our analysis, we collect multilingual text data from open-source parallel corpora. For a language \( l \), we input its corresponding text data into the model and collect text representations from the last token of each respective text. We then compute the average of these text representations to obtain the language representation \( \bm{v}_l \) for language \( l \). Finally, we calculate the similarity between two language representations as $\text{sim}(l_1, l_2) = \frac{\bm{v}_1^\top \bm{v}_2}{\Vert \bm{v}_1 \Vert  \Vert \bm{v}_2 \Vert}$. It's important to note that we extract language representations and compute similarity across each layer of the model.

\subsubsection{Analysis Setup}

We selected the Flores-200 dataset~\citep{flores} as our parallel data source, which includes 2009 sentences for each language. For the explored languages, we chose en, zh, de, fr, es, ru, it, pt, nl, pl, ja, vi, cs, tr, hu, el, sv, ro, uk, and hi, based on their highest language proportions in our pre-training data. For comparison, we also analyzed the BLOOM-7B1 model~\citep{bloom}. For this model, we considered en, zh, fr, es, ru, pt, nl, pl, ja, vi, cs, tr, hu, el, sv, ro, uk, hi, fi, and th.

\subsubsection{Results}

\Cref{similarity} illustrates the similarities distribution of multilingual representations in the intermediate layers of two models, with languages ordered according to the amount of language resources. It is apparent that for the BLOOM-7B, lower multilingual representation similarities tend to occur between the top 10 languages with higher resource availability and the bottom 10 languages with lower resource availability. In contrast, our model learn more consistent multilingual representations for all the languages we explored. This indicates that our model possesses a higher degree of multilingual balance, which is also reflected in our multilingual evaluation results and pre-training corpus.

Furthermore, we calculate the average similarity for each layer of the two models, as shown in \Cref{layersimilarity} (\Cref{sec:neuron_result}). For our model, it can be observed that there is a significant increase in similarity from the embedding layer to layer 0, reaching a very high level. As the depth of the model increases, the similarity continues to rise, indicating that the model learns richer multilingual alignment information in these layers. Subsequently, there is a sharp decrease in similarity from layer 28 to layer 29, suggesting that language-specific multilingual representations in the final layer are learned to predict the diverse multilingual vocabulary. For BLOOM-7B1, the trend of similarity changes across layers is similar, initially increasing and then decreasing, but the changes are more gradual in magnitude.

\subsection{Detailed Analysis Results}
\label{sec:neuron_result}

\begin{figure*}[t]
\centering
\centerline{\includegraphics[scale=0.3]{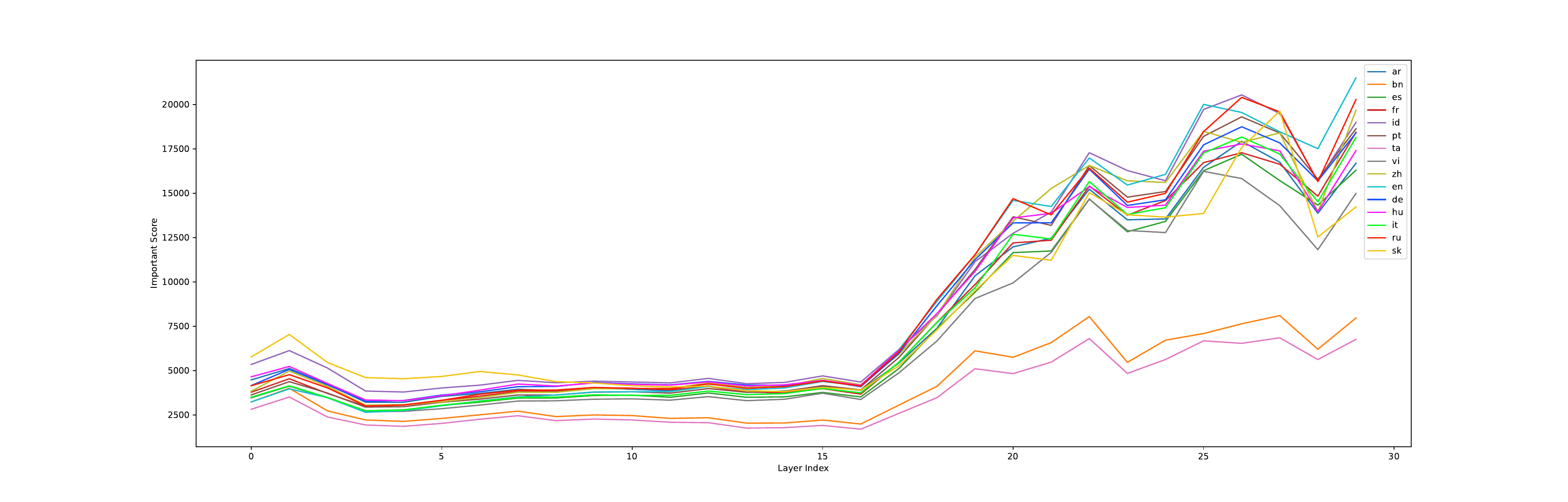}}
\caption{Importance of model layers across various language settings.}
\label{layer}
\end{figure*}

\begin{figure*}[t]
\centering
\centerline{\includegraphics[scale=0.3]{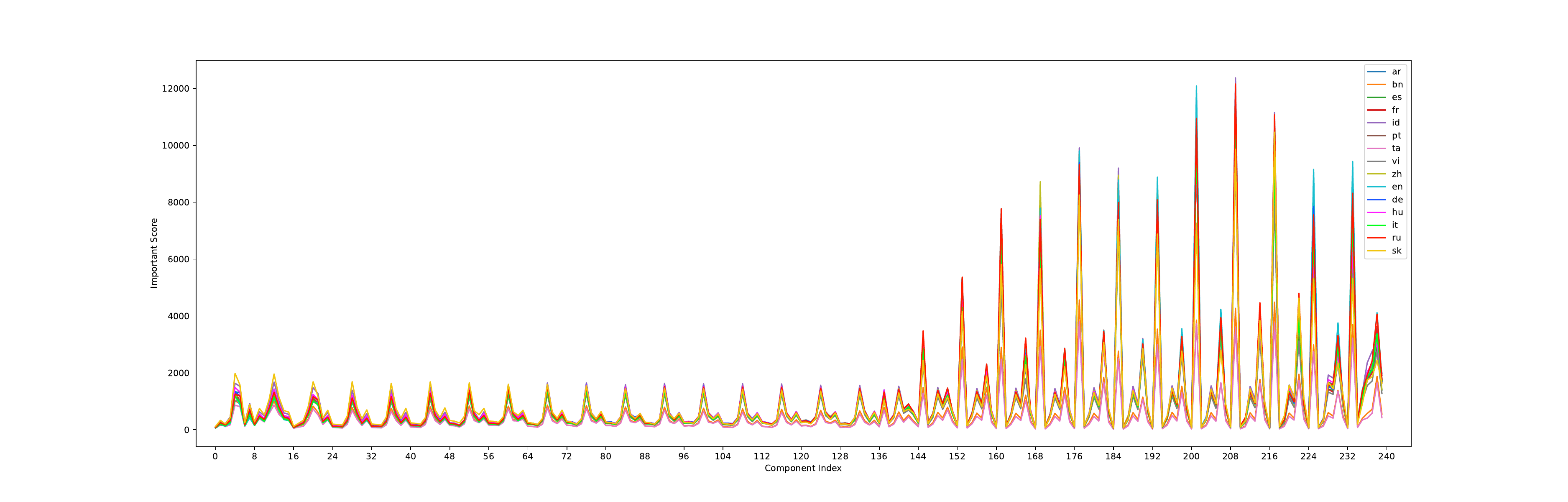}}
\caption{Importance of model components across various language settings.}
\label{component}
\end{figure*}

\begin{figure*}[ht]
\begin{center}
\includegraphics[width=0.9\linewidth]{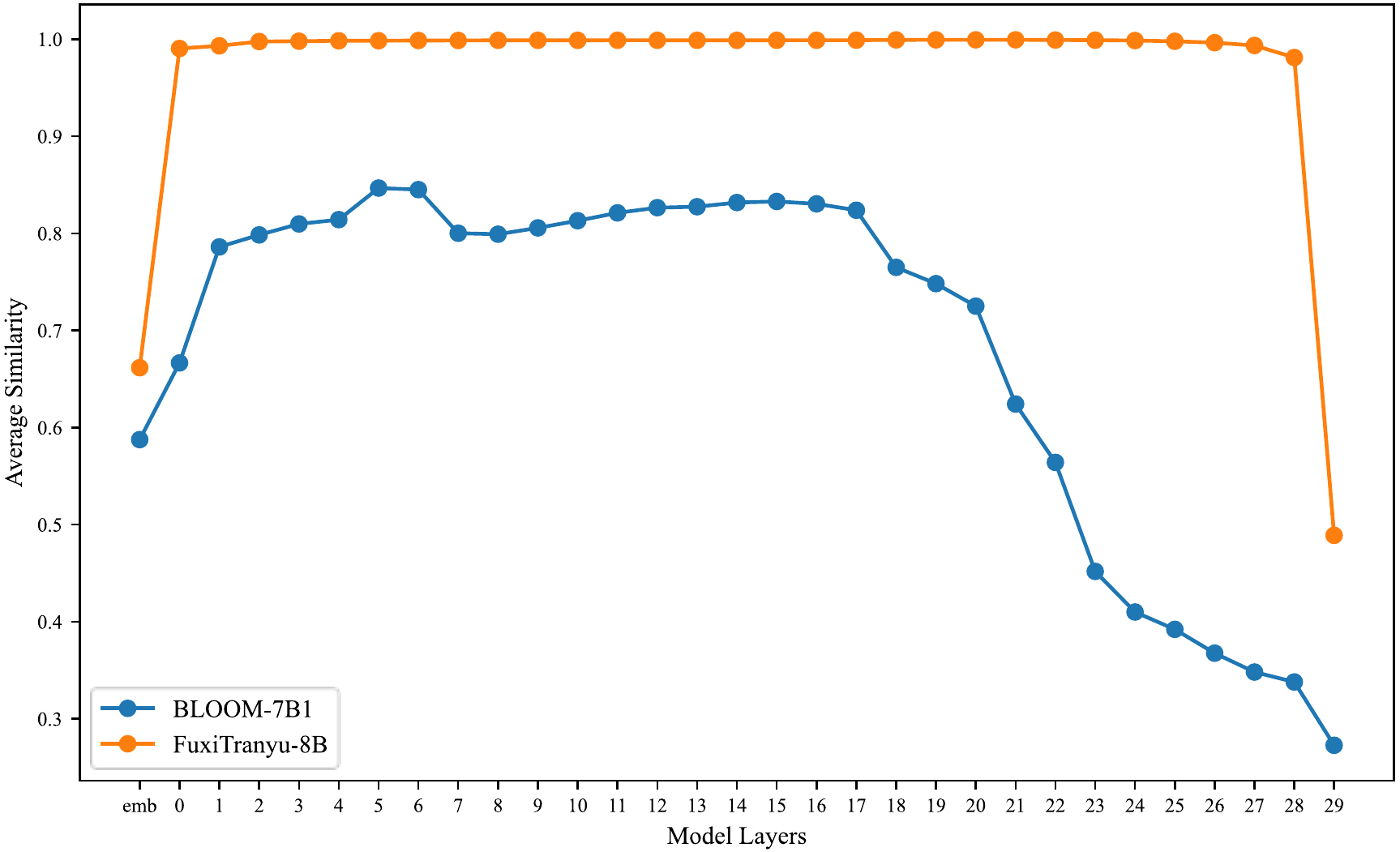} 
\caption{Averaged similarity distribution of multilingual representations for each layer of BLOOM-7B1 and FuxiTranyu-8B, with ``emb'' denoting the embedding layer.}
\label{layersimilarity}
\end{center}
\end{figure*}

We present the varying importance of different layers across diverse language inputs in \Cref{layer}. \Cref{component} shows the significance of various components across different language inputs, with 8 components per layer.
Furthermore, we calculate the average similarity of multilingual representations across model layers, as shown in \Cref{layersimilarity}.

\end{document}